\documentclass[runningheads]{llncs}

\usepackage{eccv}

\usepackage{eccvabbrv}

\usepackage{graphicx}
\usepackage{booktabs}

\usepackage{array}

\usepackage{multirow}
\usepackage{makecell}
\usepackage{color}
 \usepackage{booktabs}
\usepackage{tabularx}
\usepackage{adjustbox}
  
\usepackage[ruled,vlined]{algorithm2e}

\usepackage[accsupp]{axessibility}  

\usepackage{hyperref}

\usepackage{orcidlink}

\begin{document}

\title{CDER-SME: A Cross-Device Event-RGB Micro-Expression Dataset under Multi-Level Stress Induction} 

\titlerunning{CDER-SME}

\author{Jingting Li\inst{1,2}\orcidlink{0000-0001-8742-8488}$^{\text{\textdagger}}$ \and
Hui Sha\inst{1,3}$^{\text{\textdagger}}$\and
Su-Jing Wang\inst{1,2}$^{*}$\orcidlink{0000-0002-8774-6328}}

\authorrunning{J.~Li et al.}

\institute{State Key Laboratory of Cognitive Science and Mental Health, Institute of Psychology, Chinese Academy of Sciences, Beijing, 100101, China \and
Department of Psychology, University of the Chinese Academy of Sciences, Beijing, 100049, China \and
School of Computer Science, Jiangsu University of Science and Technology, Jiangsu, 212028, China \\
$^{\text{\textdagger}}$ Equal contribution, $^{*}$Corresponding author: \email{\{wangsujing\}@psych.ac.cn}}

\maketitle

\begin{abstract}
  Micro-expression recognition (MER) in realistic scenarios demands high temporal sensitivity and ecological validity, yet existing benchmarks are largely constrained to laboratory-controlled settings and rigid hardware-coupled sensing. We introduce CDER-SME, a cross-device Event–RGB dataset collected under a multi-level stress induction framework (cognitive and social) to elicit spontaneous emotional leakage. To enable reproducible acquisition with independent, decoupled sensors, we provide a hardware-agnostic alignment pipeline for temporal synchronization and landmark-guided spatial registration. CDER-SME adopts a three-tier structure with 92 subjects and 1,963 expert-annotated samples (Action Units and emotions), including 790 Event–RGB pairs and 210 high-fidelity aligned pairs. We further report a reproducible multimodal baseline, where cross-modal fusion improves performance over single-modality counterparts, supporting the complementarity of event dynamics and RGB cues. By removing the need for coaxial calibration, CDER-SME offers a practical benchmark for cross-device alignment and deployable Event–RGB MER in real-world affective intelligence.
\end{abstract}

\section{Introduction}
\label{sec:intro}

Emotion understanding is central to affective intelligence~\cite{picard1997affective, pantic2005affective}, where facial behavior provides a direct nonverbal channel~\cite{ekman1978facial}. Among different expression categories, micro-expressions (MEs) are spontaneous facial actions largely beyond voluntary control~\cite{ekman2009telling}. They are characterized by very short duration (around 307-327 ms), subtle intensity, and local activation confined to small facial muscle groups~\cite{li2025could}. Because MEs often occur when individuals have strong motivation to conceal or suppress genuine emotions, they offer a unique window into latent affective states, with high practical relevance to security interviews, clinical psychological assessment, and deception-related applications~\cite{ekman2009lie,porter2008reading}.
\par
Despite the recent success of deep learning and large-scale models in macro-expression (MaE) recognition, ME analysis remains highly challenging~\cite{li2022deep}, due to (i) weak and noise-sensitive cues and (ii) limited data scale. 
More importantly, the ecological validity of current datasets is limited: many samples are collected in controlled settings where participants passively watch emotion-eliciting videos~\cite{yan2014casme2,li2022cas3,zhao2024DFME}, failing to reflect the psychological antagonism induced by high-pressure concealment in real-world scenarios. In addition, the sensing modality is often overly constrained. 
Conventional RGB cameras face intrinsic limitations when capturing rapid and subtle facial motion, including restricted frame rates, motion blur, and substantial redundancy in static background content. Although prior work has explored 3D and thermal infrared modalities, the field still lacks sensors that can natively capture fine-grained facial dynamics with high temporal resolution—most notably, event-based cameras.
\par
Stress is a complex psycho-physiological response to challenging stressors and constitutes a key driver of MEs in realistic settings (e.g., deception or high task load)~\cite{dickerson2004acute}. In laboratory environments, a stress-induction paradigm, such as social-evaluative pressure and cognitive-load pressure~\cite{kirschbaum1993trier}, can more faithfully reproduce the process of emotion suppression, thereby eliciting MEs with higher ecological validity. 
Accordingly, we elicit ecologically valid ME data through a multi-level stress paradigm, incorporating a classical Stroop-based cognitive stress task~\cite{stroop1935studies} and an active deception task designed to induce social stress~\cite{kirschbaum1993trier}.
\par
Besides, to overcome the limits of RGB cameras in capturing high-speed micro-motions, we incorporate event-based cameras, neuromorphic sensors that encode only intensity changes as asynchronous event streams, offering microsecond temporal resolution and a wide dynamic range~\cite{gallego2020event}. Integrating event sensing into ME analysis reduces static redundancy while faithfully tracking the transient spatiotemporal dynamics of MEs, thereby narrowing the gap between lab settings and real-world deployment. Unlike existing DVS-RGB datasets (e.g., NEFER) that rely on coaxial or rigid hardware coupling, our CDER-SME dataset targets a more practical decoupled cross-device setting, reflecting the realistic need to augment deployed RGB systems with independent neuromorphic modules. Crucially, we show that high-fidelity affect perception remains achievable under non-pre-calibrated, non-coaxial conditions via rigorous algorithmic alignment, improving the dataset’s utility.
\par
In sum, we summarize our contributions as follows:
\begin{itemize}
    \item A high-ecological-validity stress-induction paradigm for ME collection. Unlike conventional passive stimulus protocols, we design a two-dimensional framework covering cognitive stress and social stress. With graded task intensity, our setup better approximates real-world emotion suppression under psychological pressure, improving the practical relevance of collected MEs.
    \item  A Cross-device Event–RGB ME dataset, incorporating event cameras—characterized by high temporal resolution and low latency—into ME capture, synchronized with a conventional high-resolution RGB camera. This dataset fills the gap of neuromorphic visual sensing in ME research and provides non-redundant measurements for studying highly transient, potentially non-linear facial motion patterns.
    \item Cross-device spatiotemporal alignment and multimodal benchmarks. To address spatiotemporal inconsistencies between heterogeneous sensors (event streams vs. frame images), we develop a reproducible calibration and alignment pipeline. Based on the resulting dataset, we establish micro-expression recognition (MER) benchmarks for single-modality (RGB/event) and fused-modality settings, thereby enabling fair comparison and facilitating future multimodal ME research. Crucially, by removing the need for rigid hardware coupling, our benchmark enables plug-and-play augmentation of existing RGB systems with independent neuromorphic modules, making event–RGB ME research substantially more deployable and reproducible.
\end{itemize}
Specifically, to maximize utility, CDER-SME is organized into a three-tier structure: Tier-1 comprises 1,963 RGB samples from 92 subjects for single-modality training; Tier-2 provides 790 raw Event-RGB pairs ; and Tier-3, i.e., the benchmark core, consists of 210 high-precision aligned samples from 14 subjects. 
We will release (i) the raw RGB videos and event streams in their original formats, (ii) comprehensive annotations including onset/apex/offset, AUs, laterality, emotion labels, and stress-condition labels, (iii) alignment metadata such as the estimated cross-device temporal offset, face crops, and a list of failure cases, and (iv) the full evaluation protocol together with baseline and evaluation code to facilitate reproducibility. Due to the double-blind review policy, we provide an anonymized release package in the Supp. without revealing identifying information. 

\section{Related Works}

\subsection{Micro-expression Databases}
\begin{table}[tb]
\centering
\caption{Comparison of ME databases.}
\label{tab:me_databases}
\scriptsize
\setlength{\tabcolsep}{3pt}
\renewcommand{\arraystretch}{1.15}

\begin{adjustbox}{max width=\linewidth}
\begin{tabular}{@{}l c c c c p{2.2cm} p{2.4cm} c@{}}
\toprule
\textbf{Database} &
\textbf{\#Subj.} &
\textbf{\#MEs} &
\textbf{FPS} &
\textbf{Face Res.} &
\textbf{Hardware} &
\textbf{Emotion} &
\textbf{AU} \\
\midrule

\multirow{3}{*}{SMIC~\cite{li2013smic}}
& 16 & 164 & 100 & \multirow{3}{*}{190$\times$230}
& High-speed camera
& \multirow{3}{*}{3 classes}
& \multirow{3}{*}{None} \\
& 8 & 71 & 25 &
& Near-infrared camera
& & \\
& 8 & 71 & 25 &
& Ordinary camera
& & \\
\midrule

CASME~\cite{yan2013casme}
& 19 & 195 & 60 & 150$\times$190
& RGB
& 8 classes
& 12+ \\
CASME II~\cite{yan2014casme2}
& 26 & 247 & 200 & 280$\times$340
& RGB
& 5 classes
& 11+ \\
CAS(ME)$^{2}$~\cite{qu2018cas(me)2}
& 22 & 57 & 30 & 200$\times$340
& RGB
& 4 classes
& 28 \\
SAMM~\cite{Davison2018samm}
& 32 & 159 & 200 & 400$\times$400
& Grayscale
& 7 classes
& All \\
MEVIEW~\cite{husak2017spotting}
& 16 & 31 & 30 & N/A
& RGB
& 7 classes
& 7 \\
MMEW~\cite{ben2021video}
& 36 & 300 & 90 & 400$\times$400
& RGB
& 7 classes
& 17 \\
CAS(ME)$^{3}$~\cite{li2022cas3}
& 131 & 1030 & 30 & 250$\times$300
& 3D (Depth+RGB)
& 7 / 4 classes
& 28 \\
4DME~\cite{li20234DME}
& 56 & 267 & 60 & 160$\times$200
& 4D (Gray+Depth+RGB)
& 10 classes
& 22 \\
DFME~\cite{zhao2024DFME}
& 671 & 7{,}526 & 500/300/200
& $\sim$250$\times$300
& RGB
& 7 classes
& 24 \\
CASMEMG~\cite{li2025could}
& 35 & 233 & 30 & 700$\times$900
& EMG+RGB
& 7 / 2 classes
& 27 \\
\bottomrule
\end{tabular}
\end{adjustbox}
\end{table}

\noindent\textbf{Elicitation Paradigms and Ecological Validity.}
The ecological validity of ME datasets is intrinsically tied to the induction paradigm. Early datasets, such as USF HD~\cite{shreve2011macro} and Polikovsky~\cite{polikovsky2009facial}, primarily relied on \textit{posed} expressions, which are governed by voluntary motor control and lack the spontaneous leakage characteristic of genuine MEs. To address this, the second generation of datasets (e.g., CASME series~\cite{yan2013casme,yan2014casme2,qu2018cas(me)2}, SMIC~\cite{li2013smic}, SAMM~\cite{Davison2018samm}) adopted the \textit{neutralization instruction paradigm}, where participants suppress emotions while viewing affective stimuli. While these controlled settings provide high-quality benchmarks, their ecological validity remains constrained by the laboratory environment.
Recent trends have shifted toward \textit{high ecological validity} in realistic contexts. MEVIEW~\cite{husak2017spotting} utilizes in-the-wild footage (e.g., poker games), yet suffers from uncontrolled variables like head motion and occlusion. CAS(ME)$^3$~\cite{li2022cas3} introduces a simulated crime paradigm to bridge the gap between lab control and real-world stress. However, existing high-validity data often rely on a single interrogation context with limited sample diversity. Our work extends this frontier by introducing a multi-dimensional stress-induction framework, incorporating both cognitive and social pressures. This paradigm better simulates the complex psychological mechanisms of emotion suppression in high-stakes scenarios, yielding more diverse and authentic spontaneous MEs.

\noindent\textbf{Evolution of Sensing Modalities.}
As listed in \cref{tab:me_databases}, standard ME databases are predominantly anchored in frame-based RGB acquisition. Despite efforts to increase temporal granularity (e.g., DFME~\cite{zhao2024DFME} at 500 fps), frame-based sensors are limited by the trade-off between exposure time and noise, often leading to motion blur or information loss under varying illumination. Furthermore, discrete sampling inherently fails to capture the continuous, ultra-short dynamics of muscle activations at the onset and offset phases.
While multimodal extensions like CAS(ME)$^3$ (Depth), 4DME~\cite{li20234DME} (3D), and CASMEMG~\cite{li2025could} (sEMG) provide geometric or physiological insights, they remain ``frame-homologous'' or lack pixel-level spatial-temporal precision. \textbf{Event-based vision} offers a paradigm shift; unlike global-shutter frames, event cameras sense per-pixel intensity changes asynchronously. This neuromorphic approach provides microsecond-level resolution and high dynamic range, making it uniquely suited for capturing the transient, localized edge motions of MEs. Our work establishes the first benchmark that leverages real-world synchronized Event-RGB streams to eliminate the modality-level blind spot in current ME research.

\subsection{Event-based Facial Analysis and MER Methods}

\noindent\textbf{Neuromorphic Facial Datasets.}
Research into event-based facial analysis is in its nascent stage. NEFER~\cite{berlincioni2023neuromorphic} provided the first paired RGB-event stream for general expression recognition, but lacked the subtle Action Unit (AU) granularity required for MEs. Other datasets like FACEMORPHIC~\cite{becattini2024neuromorphic} and VETEX~\cite{adra2024beyond} focus on posed expressions or limited induced AUs. A significant portion of current event-based ME research still relies on \textit{simulated} data (e.g., via DVS-Voltmeter), which fails to model the non-ideal noise and latency of physical hardware. By providing a hardware-synchronized dataset of spontaneous MEs, we provide a more robust foundation for training neuromorphic models.

\noindent\textbf{Micro-expression Recognition Approaches.}
Traditional MER methods primarily optimize for frame-based sequences using CNNs or Vision Transformers. The introduction of event streams necessitates specialized architectures to handle asynchronous data. Methods like GLEFFN~\cite{guo2023gleffn} convert events into Local Counting Images (LCI), while ESTME~\cite{xiao2024estme} focuses on spatio-temporal motion enhancement. However, effectively fusing the high-frequency motion cues from events with the rich textural information of RGB remains a challenge. Current SNN-based approaches~\cite{mastropasqua2025exploring} offer promise but struggle with convergence stability. Our work addresses these gaps by establishing a multimodal benchmark and a spatiotemporal alignment pipeline, facilitating future research into cross-modal fusion and interpretability in ME analysis.

\section{Database Construction}
To bridge the gap between lab-controlled recordings and real-world emotional leakage, we propose a multi-dimensional stress induction framework mapping ``Stressor-Response'' theory into a controllable design. By manipulating cognitive load and social evaluative threat, we elicit ecologically valid MEs in high-stakes scenarios.
All experimental protocols and data collection procedures were approved by the Institutional Review Board. Prior to the study, written informed consent was obtained from all 92 participants, who explicitly granted permission for their facial data to be utilized for scientific research. 

\subsection{Multi-Level Stress Induction Paradigm}

\subsubsection{A. Cognitive Stress Module.}
This module induces stress via information-processing conflict and task load using a modified Stroop task, in which graded stress levels are achieved by systematically varying the degree of conflict, time constraints, and task-rule complexity.

\begin{figure}[tb]
    \centering
    \includegraphics[width=0.8\linewidth]{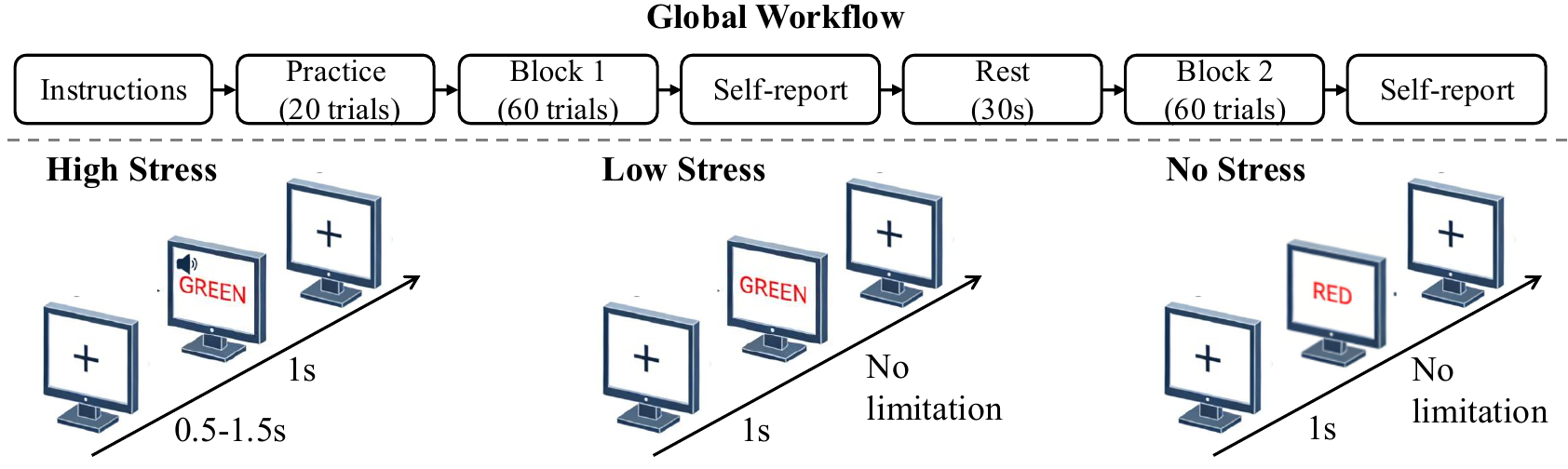}
    \caption{Workflow for cognitive stress Module}
    \label{fig:stroop_process}
\end{figure}

\textbf{Design \& Procedure.} We employed a single-factor between-subject design with three levels ($N=92$ based on G-power calculation): \textit{High}, \textit{Low}, and \textit{No Stress}. As illustrated in \cref{fig:stroop_process}, the formal experiment comprised two blocks (120 trials in total). Perceived stress was rated after each block on a 5-point Likert scale to validate the induction (1: completely relaxed, 5: very tense).

\par
\textbf{Stress Manipulation.} 
(1) \textit{High Stress}: A dual-conflict Stroop task was used. Participants responded to the ink color of a visual word while an asynchronous audio color cue was played as interference. Each trial started with a random fixation (0.5--1.5s), followed by a 1s stimulus. Responses exceeding the 1s window were marked as errors, creating high time pressure. 
(2) \textit{Low Stress}: A standard congruency task where the stimulus remained until a response was recorded, removing time-constrained pressure. 
(3) \textit{No Stress}: Only two congruent colors (red/green) were used as a baseline.
\par
\textbf{Rationale for MEs.} This paradigm systematically induces cognitive overload, leading to micro-fluctuations in the periocular and brow regions, and requiring high-temporal event sensing to capture dynamics which might be blurred in 30fps sequences.

\subsubsection{B. Social Stress Module.}
The social stress paradigm simulates stress experiences under socially evaluative conditions. We adopt a modified speech task and construct three stress conditions (reading aloud / public speaking / deceptive speech). The intensity of social stress is progressively enhanced through mechanisms such as social evaluation, being observed, deception-related risk, and potential consequences.

\textbf{Design \& Topics.} A single-factor within-subject design ($N=45$ based on G-power calculation) was adopted, involving three counterbalanced conditions: Deceptive Speech, Standard Speech, and Reading. 
Randomized topics included University Life, Childhood, and Best Friend. In each condition, participants prepared their content in advance. Following each session, the same 5-point Likert scale was used for self-reported stress. 

\textbf{Stress Manipulation.} 
(1) \textit{High Stress (Deception)}: Participants were instructed to fabricate a completely false story based on the assigned topic. To maximize social pressure, they were told that: (a) two graduate-level experts would judge their honesty in real-time; (b) the session was being recorded for "AI-based lie detection"; and (c) their final compensation would be docked if they were judged as lying. 
(2) \textit{Low Stress}: A truthful speech delivered to a live audience (2--3 members). 
(3) \textit{No Stress}: Reading a neutral text with the same audience present. 
Post-experiment, participants provided a qualitative self-evaluation of their "lying performance" to ensure task engagement.

\textbf{Rationale for MEs.} The social stress module is designed to trigger "emotional leakage" when participants attempt to suppress their true feelings under the scrutiny of an audience. This results in a rich variety of suppressive MEs and atypical facial AU combinations.

\subsection{Data Acquisition Setup}
We utilized a 7th-gen event sensor ($640\!\times\!480$, microsecond-level capture capacity) and a Sony Alpha 7S III RGB camera ($3840\!\times\!2160$, 30 fps). A customized rig ensured near-identical viewpoints, with an external chronometer for temporal alignment. Soft-box lighting ($5500$K) and a green backdrop minimized event noise. The collection Environment photo is presented in Supp.

\subsection{Data Annotation}
As expressions under high-stakes pressure are inherently fragmented and suppressive, subjective self-reports or task-dependent labels fail to capture authentic facial muscle dynamics. Consequently, we adopt AU annotation and rule-based emotion derivation, the established standard in the MER community.
\par
\noindent\textbf{AU Coding and Reliability.}
Facial AUs and temporal boundaries were independently coded by three certified experts. Following the protocol in CAS(ME)$^2$~\cite{qu2018cas(me)2}, inter-rater reliability (IRR) was quantified by the consistency score: $R = \frac{2 \times A_{1,2}}{A_1 + A_2}$, where $A_{1,2}$ denotes the number of overlapping AUs between two coders.
Specifically, inter-rater reliability (IRR) reached \textbf{82.5\%} (Cognitive) and \textbf{93.1\%} (Social). Beyond standard AU categories, we further annotated the \textbf{laterality} (unilateral or bilateral) of each muscle activation to support fine-grained dynamic analysis and cross-modal learning, achieving an agreement of \textbf{99.8\%} and \textbf{97.3\%} respectively. Discrepancies were resolved via consensus arbitration.

\par
\noindent\textbf{Rule-based Emotion Derivation.}
\label{sec:au_label}
In traditional stimulus-evoked paradigms, emotion labels are typically derived based on the valence of video clips and participants' self-reports. However, in our multi-dimensional stress-induction framework, facial expressions often manifest as spontaneous and fragmented "leakages" under psychological pressure. In such high-stakes scenarios, participants' subjective ratings are frequently biased by their coping strategies.
To ensure objectivity, we derived emotion categories from AU combinations based on the FACS manual~\cite{ekman1978facial} and the established AU-ME objective coding protocol~\cite{davison2018objective}. Specifically, we standardized the AU sets by stripping laterality prefixes and implemented a \textbf{priority-based hierarchical matching mechanism} to handle blended expressions. The mapping follows a descending priority order~\cite{dong2022spontaneous}:
(i)~\textbf{Happiness}: $\{12, 6, 28\} \cap \mathcal{A} \neq \emptyset$;
(ii)~\textbf{Surprise}: $2 \in \mathcal{A}$;
(iii)~\textbf{Anger}: $\{16, 22, 23\} \cap \mathcal{A} \neq \emptyset$;
(iv)~\textbf{Fear}: $\{1, 4, 5, 25\} \subseteq \mathcal{A}$;
(v)~\textbf{Disgust}: $\{7, 24\} \cap \mathcal{A} \neq \emptyset$; and
(vi)~\textbf{Sadness}: $(\{4, 5\} \subseteq \mathcal{A}) \lor (43 \in \mathcal{A})$.
Samples that did not satisfy any prototypical rules were categorized as \textit{others}. This procedure ensures that our labels are grounded in observable facial dynamics rather than subjective inference. Please see detailed implementation of our mapping algorithm in Supp.

\subsection{Dataset Statistics \& Quality Analysis}

\noindent\textbf{Manipulation Check - Subjective Stress Efficacy.}
Subjective reports (\cref{tab:subjective_stress}) validated induction efficacy. In the Stroop task, perceived stress scores were significantly distinct across conditions: high ($M=3.65$), low ($M=2.43$), and no-stress ($M=1.69$). A one-way ANOVA indicated a significant main effect ($F(2, 177) = 89.75, p < 0.001, \eta_{p}^{2} = 0.50$), confirming successful cognitive load modulation. Similarly, in the social stress module, deceptive speech elicited significantly higher stress ($M=4.40$) compared to standard speech ($M=3.18$) and reading ($M=2.04$), confirmed by a repeated-measures ANOVA ($F(2, 132) = 63.76, p < 0.001, \eta_{p}^{2} = 0.49$). These results establish a solid ground truth for analyzing stress-driven facial dynamics.

\begin{table}[tb]
\centering
\caption{Summary of subjective stress ratings and expression counts across experimental conditions. ($M$: Mean, $SD$: Standard Deviation, $N$: Total Count).}
\label{tab:subjective_stress}
\begin{tabular}{l|ccc|cc}
\hline
\textbf{Condition} & \textbf{$M \pm SD$} & \textbf{$F$-value} & \textbf{$p$-value} & \textbf{MEs ($N$)} & \textbf{MaEs ($N$)} \\ \hline
\textit{Stroop (Cognitive)} & & & & & \\
~~High Stress & $3.65 \pm 0.86$ & \multirow{3}{*}{89.745} & \multirow{3}{*}{$<.001$} & 117 & 143 \\
~~Low Stress & $2.43 \pm 0.84$ & & & 59 & 80 \\
~~No Stress & $1.69 \pm 0.74$ & & & 52 & 98 \\ \hline
\textit{Social (Deception)} & & & & & \\
~~Deceptive Speech & $4.40 \pm 0.75$ & \multirow{3}{*}{63.761} & \multirow{3}{*}{$<.001$} & 377 & 247 \\
~~Standard Speech & $3.18 \pm 1.13$ & & & 288 & 264 \\
~~Reading & $2.04 \pm 1.04$ & & & 183 & 55 \\ \hline
\textbf{Total} & --- & --- & --- & \textbf{1,076} & \textbf{887} \\ \hline
\end{tabular}%

\end{table}

\par
\noindent\textbf{Expression Distribution and Frequency.}
Stress-expression relationships varied by module. For cognitive stress, while subjective intensity varied, ME/MaE counts showed no statistical differences across levels ($\chi^2 = 29.351, p = 0.395$ for MEs), suggesting that individual regulation thresholds may mask overt frequency changes. Conversely, social stress yielded a robust increase in emotional leakage: deceptive speech produced significantly more MEs (377) and MaEs (247) compared to reading (183 MEs, 55 MaEs). This underscores the high ecological validity of the social stress module for inducing "emotional leakage," as participants struggle to suppress true feelings under high-stakes social scrutiny.

\par
\begin{figure}[tb]
  \centering
  \includegraphics[width=\linewidth]{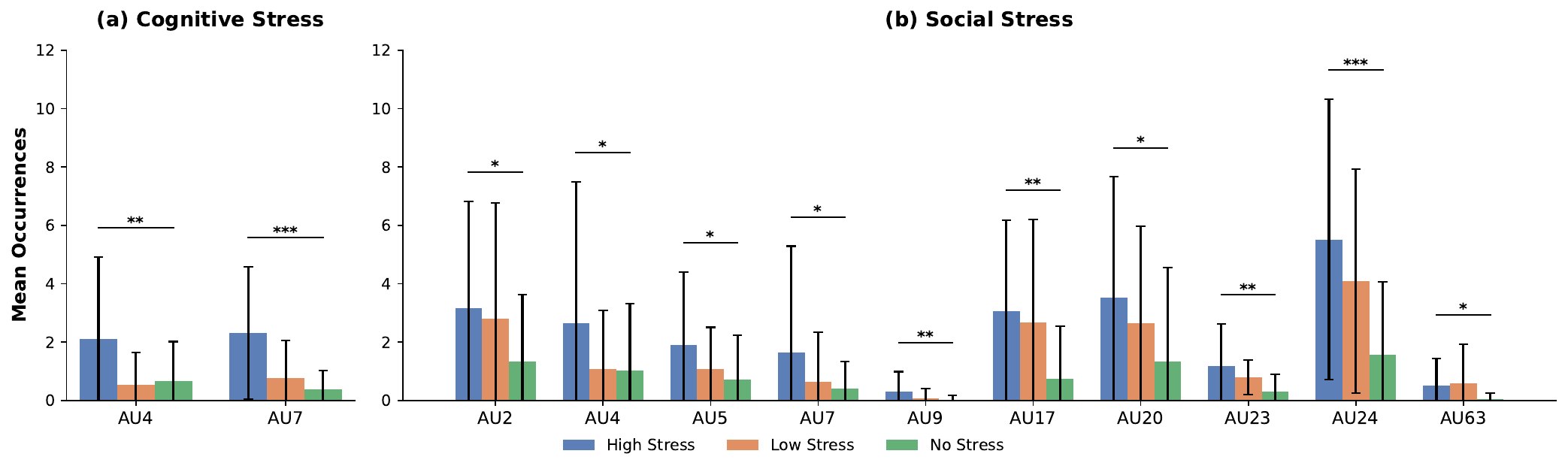}
  \caption{Significant AU occurrences under two multi-level stress induction paradigm}
  \label{fig:AU_stat}
\end{figure}

\noindent\textbf{AU Signatures under Stress.}
AU patterns provided a granular view of stress-driven facial dynamics (\cref{fig:AU_stat}).
Specifically, regarding cognitive sensitivity, brow lowering (AU4, $F=6.340, p < 0.01$) and orbital tightening (AU7, $F=13.543, p < 0.001$) served as indicators, aligning with upper facial muscle susceptibility to mental load; and regarding social sensitivity, high stress triggered broader involvement across ten AUs, including eye (AU2, 4, 5, 7, 9) and mouth (AU20, 23, 24, 63) regions. The ecological validity of CDER-SME is evidenced by unique stress-driven AU signatures. Specifically, during the Deception task, we observed a significantly higher frequency of AU24 (lip pressor). This aligns with psychological theories of ``leakage'' where subjects physically suppress speech-related anxiety under high social scrutiny, a dynamic largely absent in passive stimulus-based datasets.

\section{Cross-Device Spatial and Temporal Alignment}
\begin{figure}[tb]
    \centering
    \includegraphics[width=\linewidth]{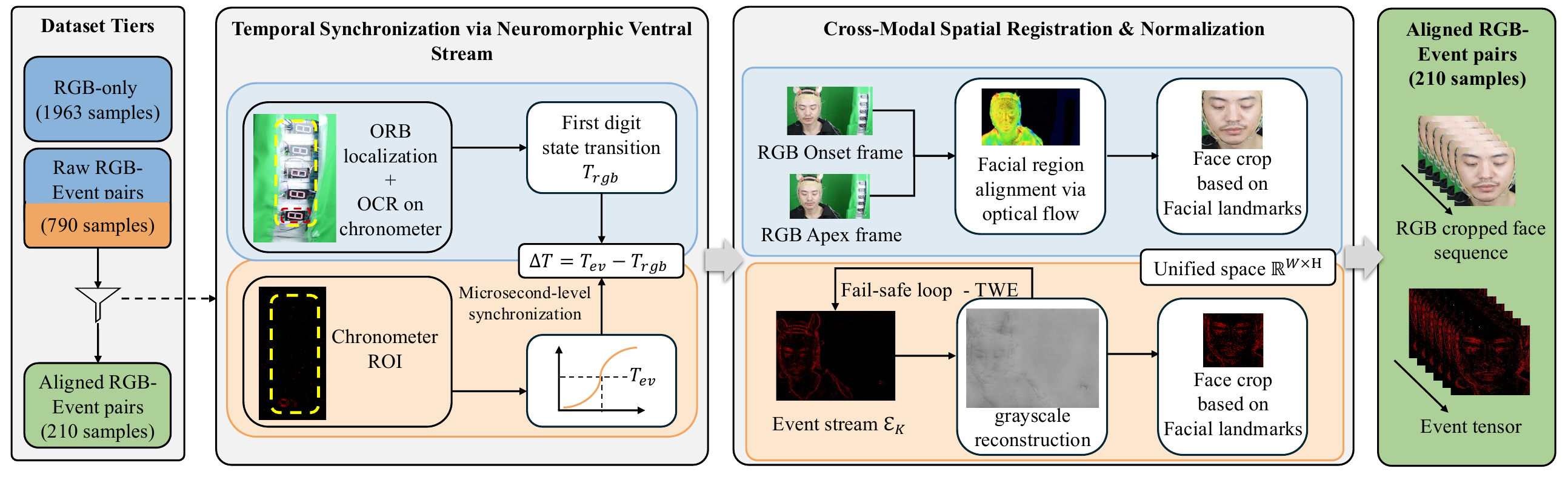}
    \caption{The proposed cross-device alignment pipeline}
    \label{fig:alignment_pipeline}
\end{figure}

While our multi-pressure paradigm yielded a total of 1,963 spontaneous expression samples in the RGB modality, hardware constraints limited the final dataset to 790 samples with concurrent RGB-Event recordings. 
Meantime, high-fidelity cross-modal ME analysis requires reliable temporal synchronization and spatial correspondence. To reflect real deployment, our sensors are decoupled and not pre-calibrated per session; we therefore design a reproducible multi-stage alignment and filtering pipeline (\cref{fig:alignment_pipeline}), yielding 210 high-precision pairs from 790 raw recordings.
%

\subsection{Chronometer-based Cross-device Temporal Synchronization}

To eliminate jitters and human error inherent in manual alignment, we implement an automated synchronization protocol based on a digital chronometer.
\par
\noindent\textbf{Event-based Change Detection.}
In the chronometer ROI $\Omega$, the event stream within $\Omega$ is modeled as a sequence of tuples $e_i = (x_i, y_i, t_i, p_i)$. The neuromorphic trigger $T_{ev}$ is determined by the cumulative density of polarity-positive events:
\begin{equation}
    \Gamma(t) = \int_{t_0}^{t} \sum_{(x,y) \in \Omega} [p_i = +1] \, dt, \quad T_{ev} = \min \{ t \mid \Gamma(t) \ge \Theta \}
\end{equation}
where $\Theta$ is a structural threshold designed to filter out sensor noise.
\par
\noindent\textbf{RGB-based Character Consistency.}
For the RGB stream, we employ ORB (Oriented FAST and Rotated BRIEF) features to localize the chronometer and apply an Optical Character Recognition (OCR) engine $\Phi$. The RGB trigger $T_{rgb}$ is the timestamp of the first detected state transition in the digit sequence:
\begin{equation}
    T_{rgb} = \min \{ t \mid \Phi(\mathcal{I}_{rgb}(t)) \neq \Phi(\mathcal{I}_{rgb}(t-\Delta t)) \}
\end{equation}
The temporal offset is thus $\Delta T = T_{ev} - T_{rgb}$, enabling temporal synchronization.

\subsection{Cross-Modal Spatial Registration and Normalization}
Spatial alignment between intensity frames $\mathcal{I}_{rgb}$ and sparse event streams $\mathcal{E}$ is non-trivial due to their heterogeneous nature and differing viewpoints. Rather than pursuing rigid pixel-to-pixel registration, which is often susceptible to parity errors and sensor noise, we adopt a Region-centric Alignment strategy to normalize modalities into a semantically consistent space.
\par
\noindent\textbf{Head Motion Suppression in RGB Modality.}
To isolate subtle MEs from significant head movements, we utilize the Dual TV-L1 optical flow to estimate the motion field $u$. To isolate MEs, we subtract the global displacement (average velocity of the stable nasal region) from the flow field.~\cite{he2022micro}. Subsequently, facial landmarks are detected to define the primary face region.
\par
\noindent\textbf{Event-to-Intensity Translation and Unified Cropping.}
For the event stream, we first transform the asynchronous data into a grid-based representation via a High Dynamic Range reconstruction paradigm~\cite{rebecq2019high}. For an event segment $\mathcal{E}_k$, the latent intensity image $\hat{\mathcal{I}}_{ev}$ is recovered by:
\begin{equation}
    \ln \hat{\mathcal{I}}_{ev}(x, t_k) = \ln \hat{\mathcal{I}}_{ev}(x, t_{k-1}) + \sum_{i} p_i C
\end{equation}
where $C$ is the contrast sensitivity. To ensure robust localization, we implement Temporal Window Expansion (TWE) to symmetrically expand the integration window ($\tilde{\mathcal{E}}_k = [t_{start}-0.5s, t_{end}+0.5s]$) if initial face detection fails. 
\par
\noindent\textbf{Region-based Cross-Modal Normalization.}
The core of our spatial alignment lies in Consistent Semantic Cropping. We apply identical geometric constraints derived from the respective landmark detectors to both $\mathcal{I}_{rgb}$ and $\hat{\mathcal{I}}_{ev}$. Both modalities are cropped and resized into a unified coordinate space $\mathbb{R}^{W \times H}$. This ensures that while the sensors may have different intrinsics, the resulting RGB image and event tensors are semantically homographic, focusing exclusively on the active facial regions required for downstream ME analysis.

\subsection{Alignment Quality Assessment}
\noindent\textbf{Quality Check.}
To ensure benchmark integrity under real sensor noise, we filter raw pairs using strict synchronization and registration criteria: from 790 recordings, 580 are excluded due to overexposure-induced temporal errors (n=270), marker occlusion/offset (n=121), and acquisition failures (n=189)(See Supp. for a detailed list). The remaining 210 pairs form a high-fidelity “gold standard”, while the full 790-pair set is released as a challenging testbed for future robust alignment research.
\par
\noindent\textbf{Representativeness Analysis.} 
While the alignment process results in a subset of 210 high-quality samples, this represents the most physically synchronized and real cross-modal micro-expression data with AU and emotion annotation currently available. To address potential concerns regarding sample size, we conducted $\chi^2$ tests (See detailed results in Supp.) which demonstrate that this subset maintains strict statistical consistency with the larger 790-sample pool in terms of expression category (micro vs. macro) distribution ($p=0.163$) and temporal duration ($p=0.051$). These results ensure that the 210-sample subset is a behaviorally representative foundation for high-precision multimodal research.

\newcommand{\mySubFigWidth}{0.11\linewidth} 

\newcommand{\imgPathPrefix}{./spatial_quality/}
\newcommand{\imgSuffix}{.jpg} 

\begin{figure}[tb]
  \centering
  
  \begin{subfigure}{\mySubFigWidth}
    \includegraphics[width=\linewidth]{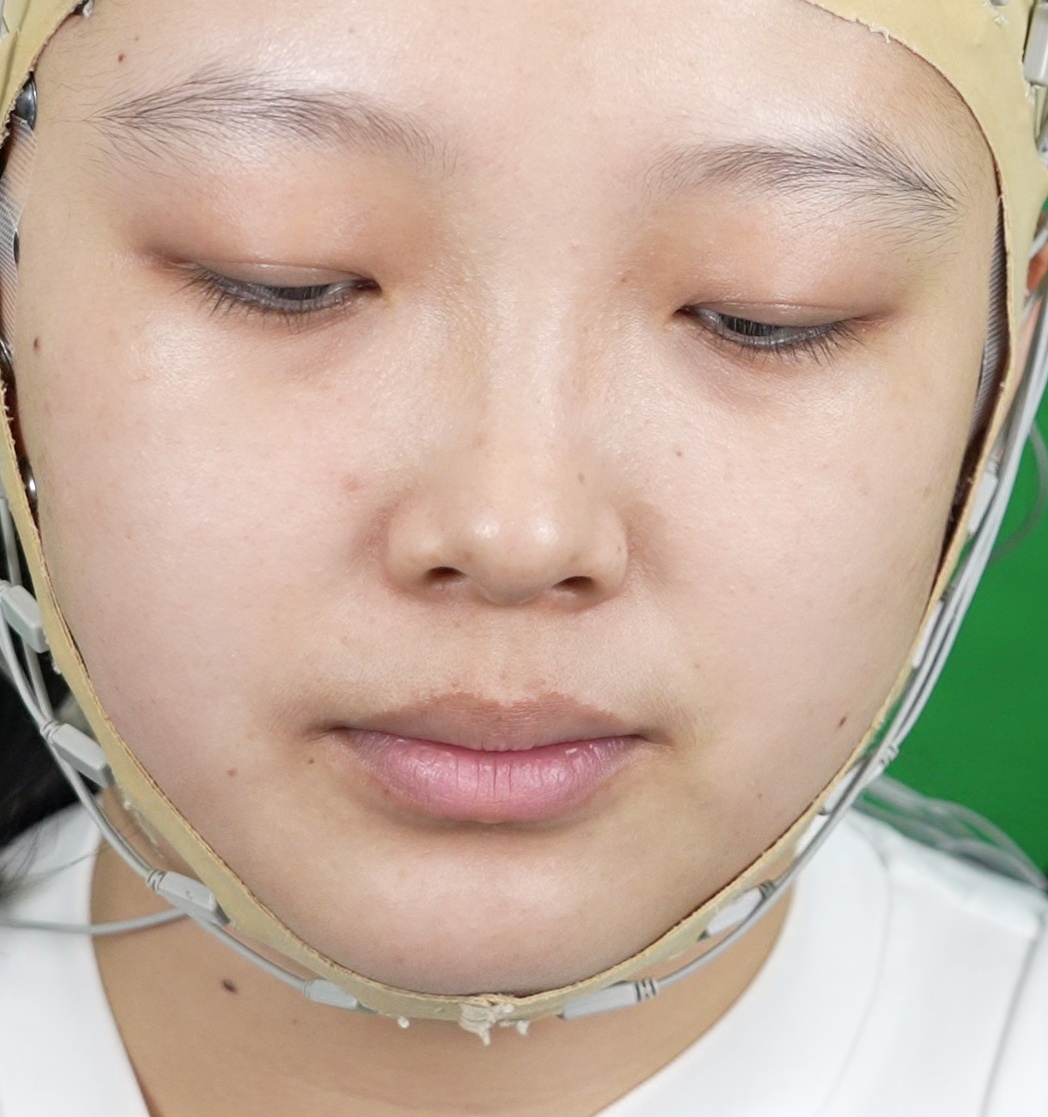}
    \label{fig:rgb-1}
  \end{subfigure}\hfill
  \begin{subfigure}{\mySubFigWidth}
    \includegraphics[width=\linewidth]{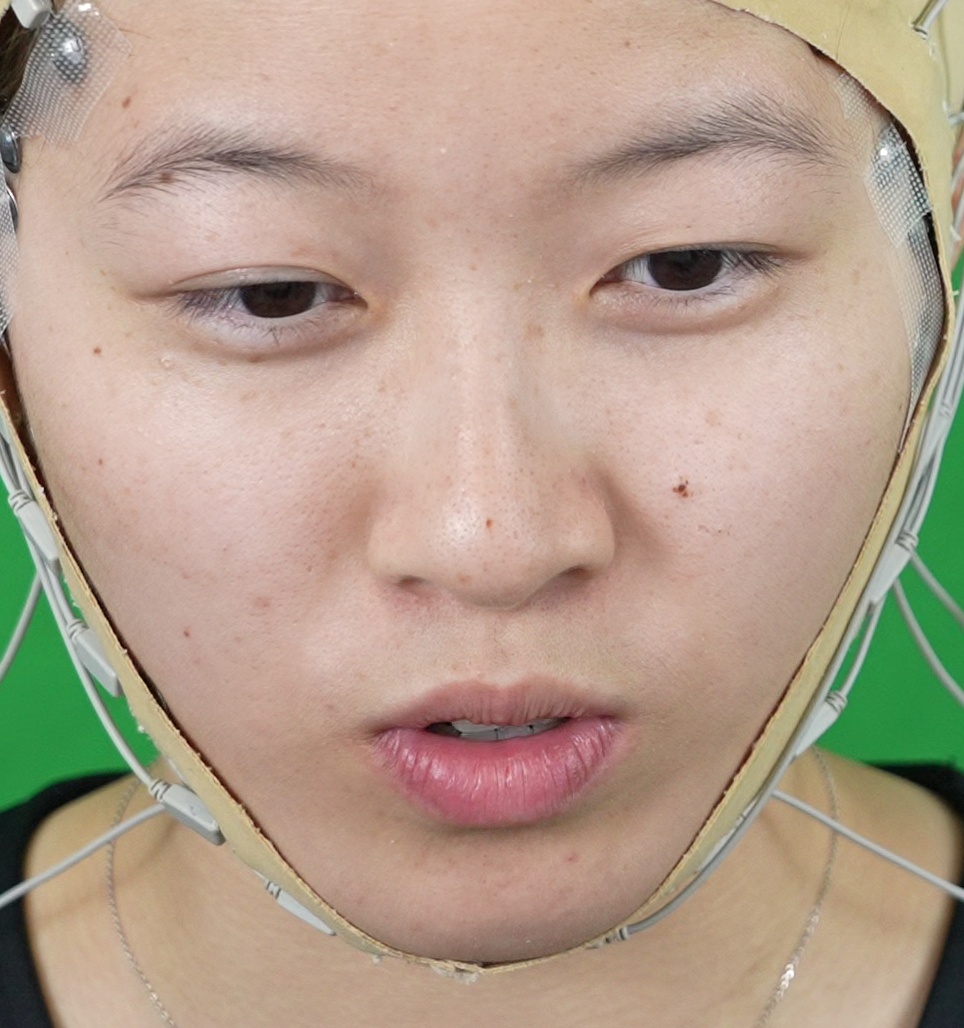}
    \label{fig:rgb-2}
  \end{subfigure}\hfill
  \begin{subfigure}{\mySubFigWidth}
    \includegraphics[width=\linewidth]{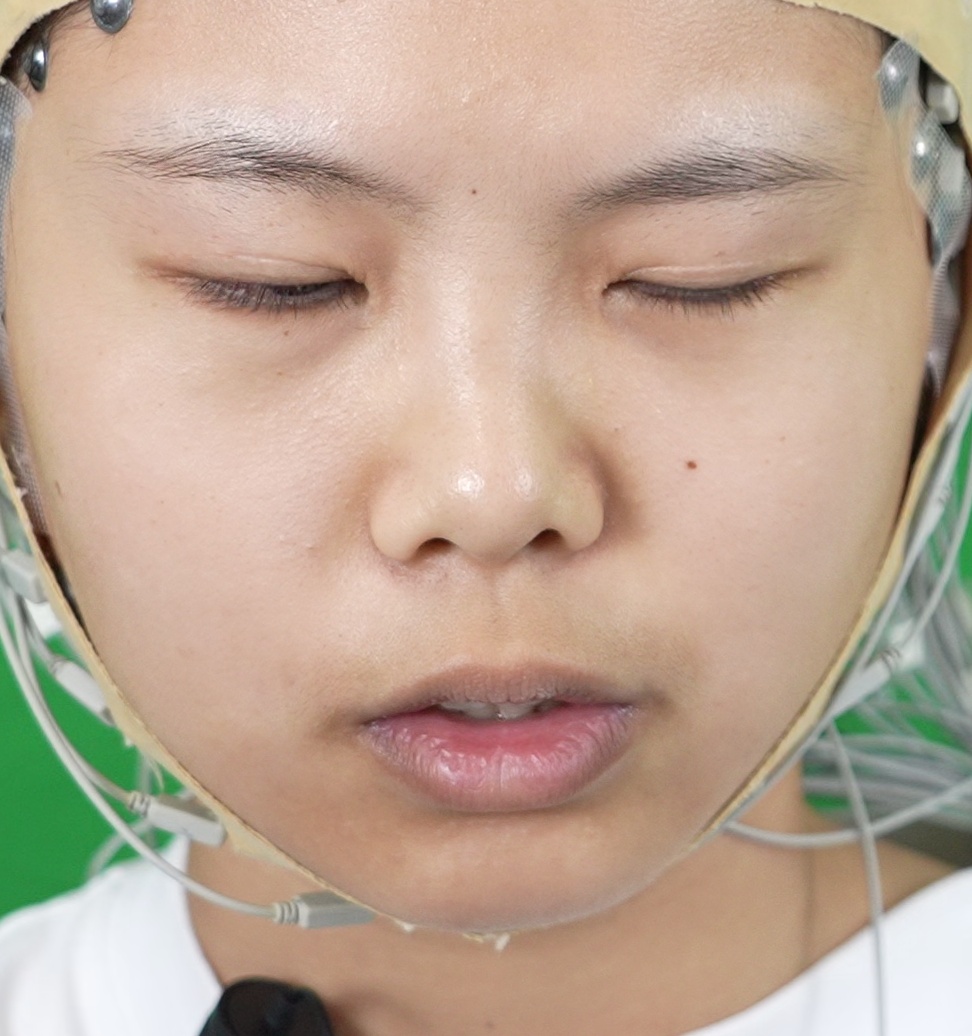}
    \label{fig:rgb-3}
  \end{subfigure}\hfill
  \begin{subfigure}{\mySubFigWidth}
    \includegraphics[width=\linewidth]{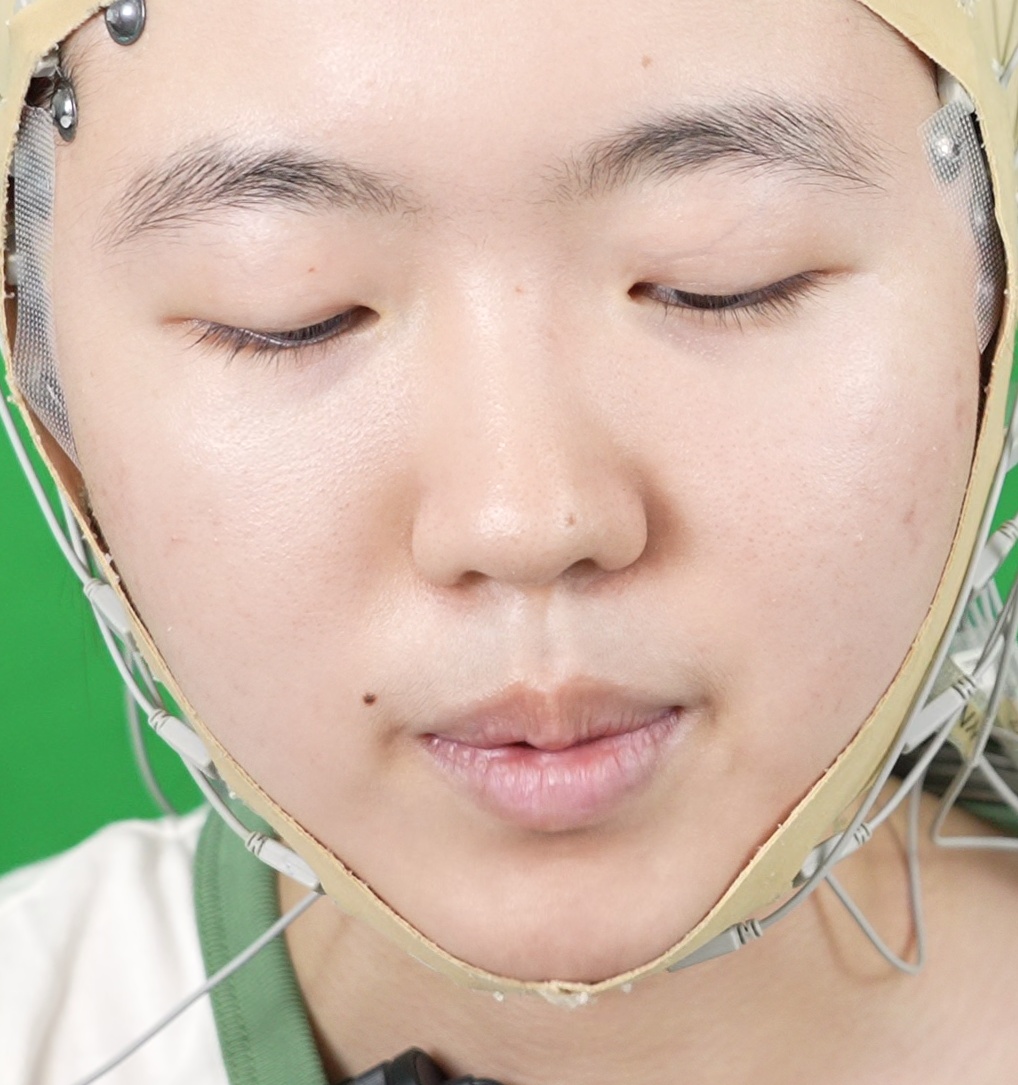}
    \label{fig:rgb-4}
  \end{subfigure}\hfill
  \begin{subfigure}{\mySubFigWidth}
    \includegraphics[width=\linewidth]{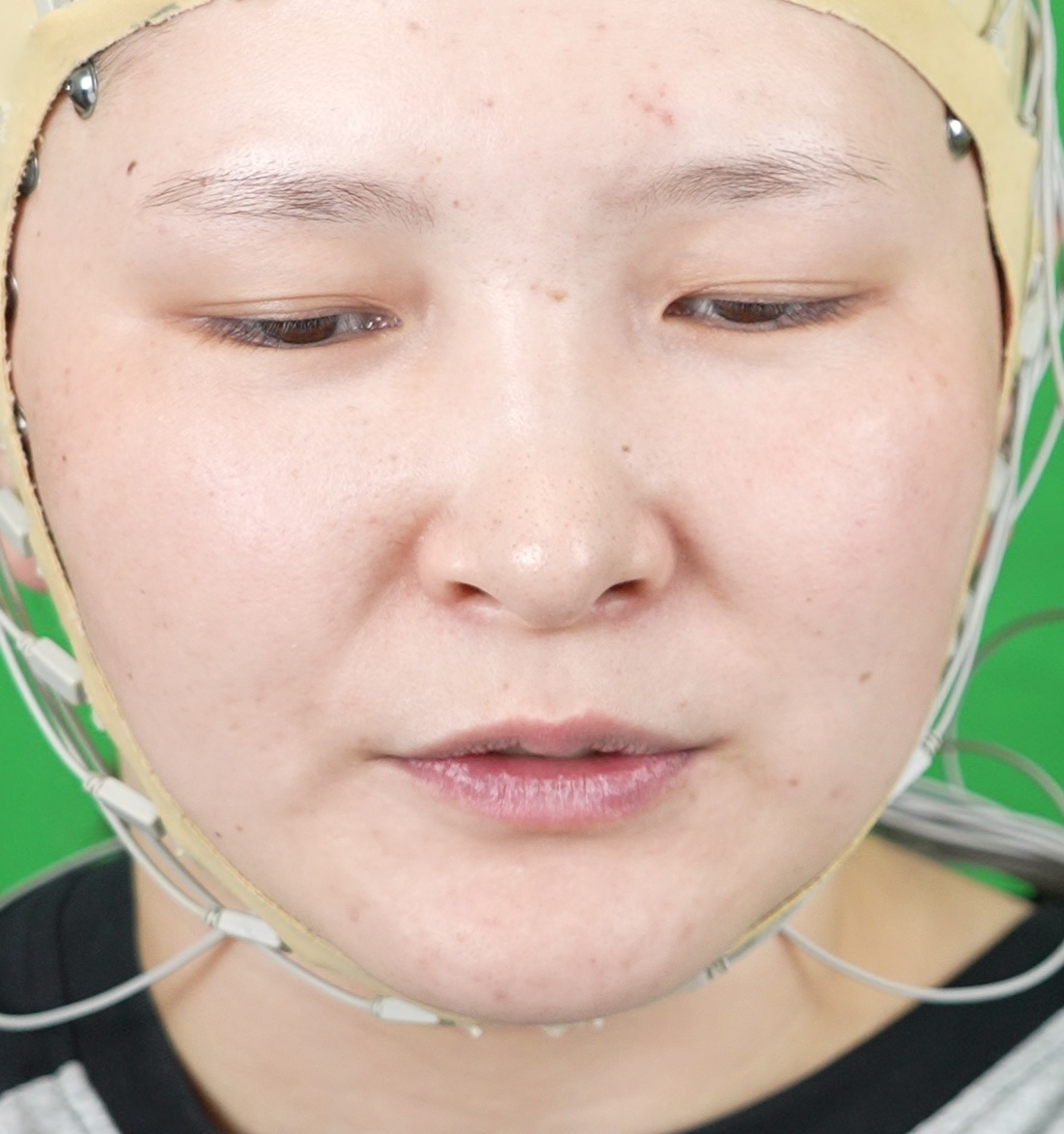}
    \label{fig:rgb-5}
  \end{subfigure}\hfill
  \begin{subfigure}{\mySubFigWidth}
    \includegraphics[width=\linewidth]{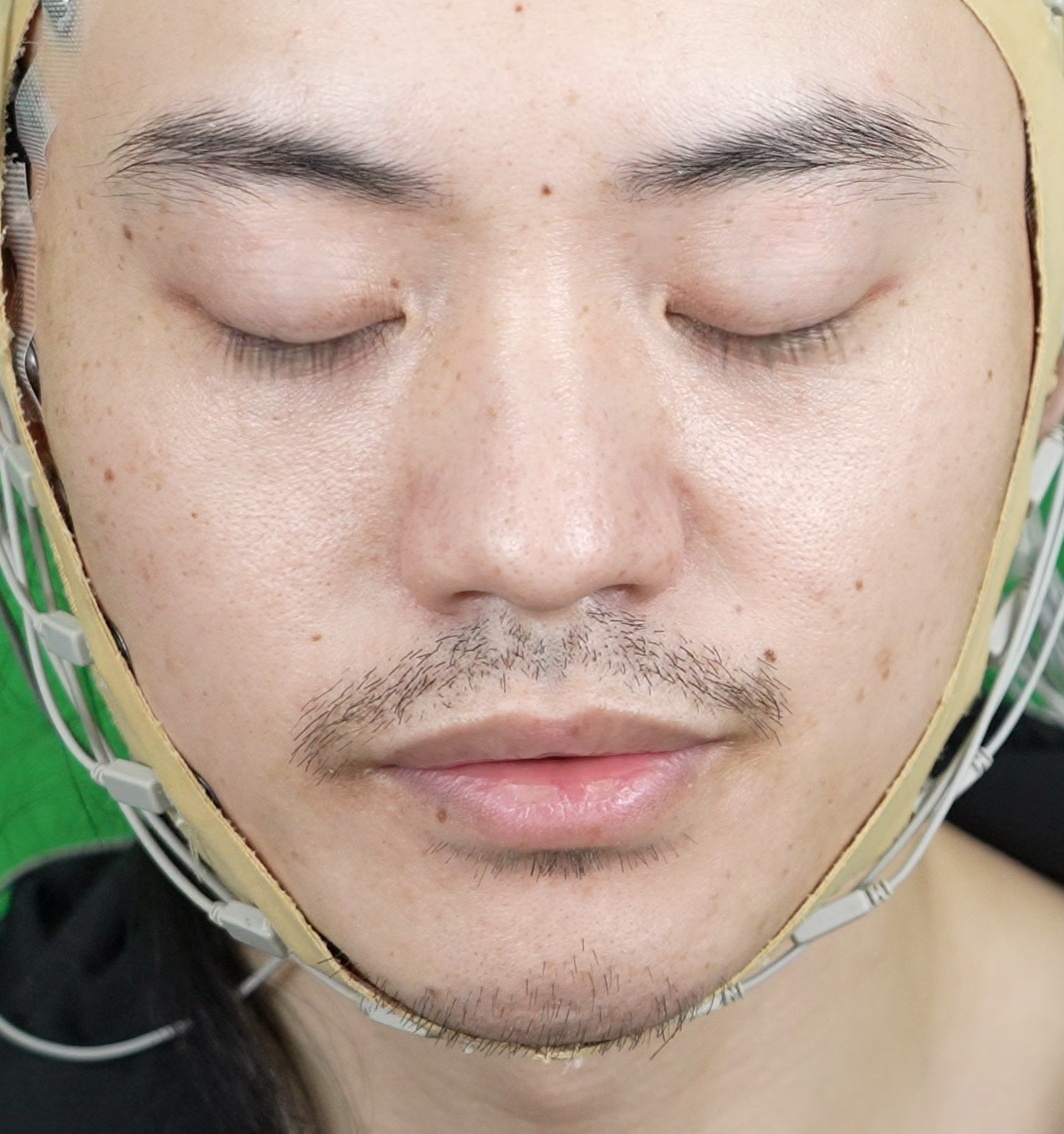}
    \label{fig:rgb-6}
  \end{subfigure}\hfill
  \begin{subfigure}{\mySubFigWidth}
    \includegraphics[width=\linewidth]{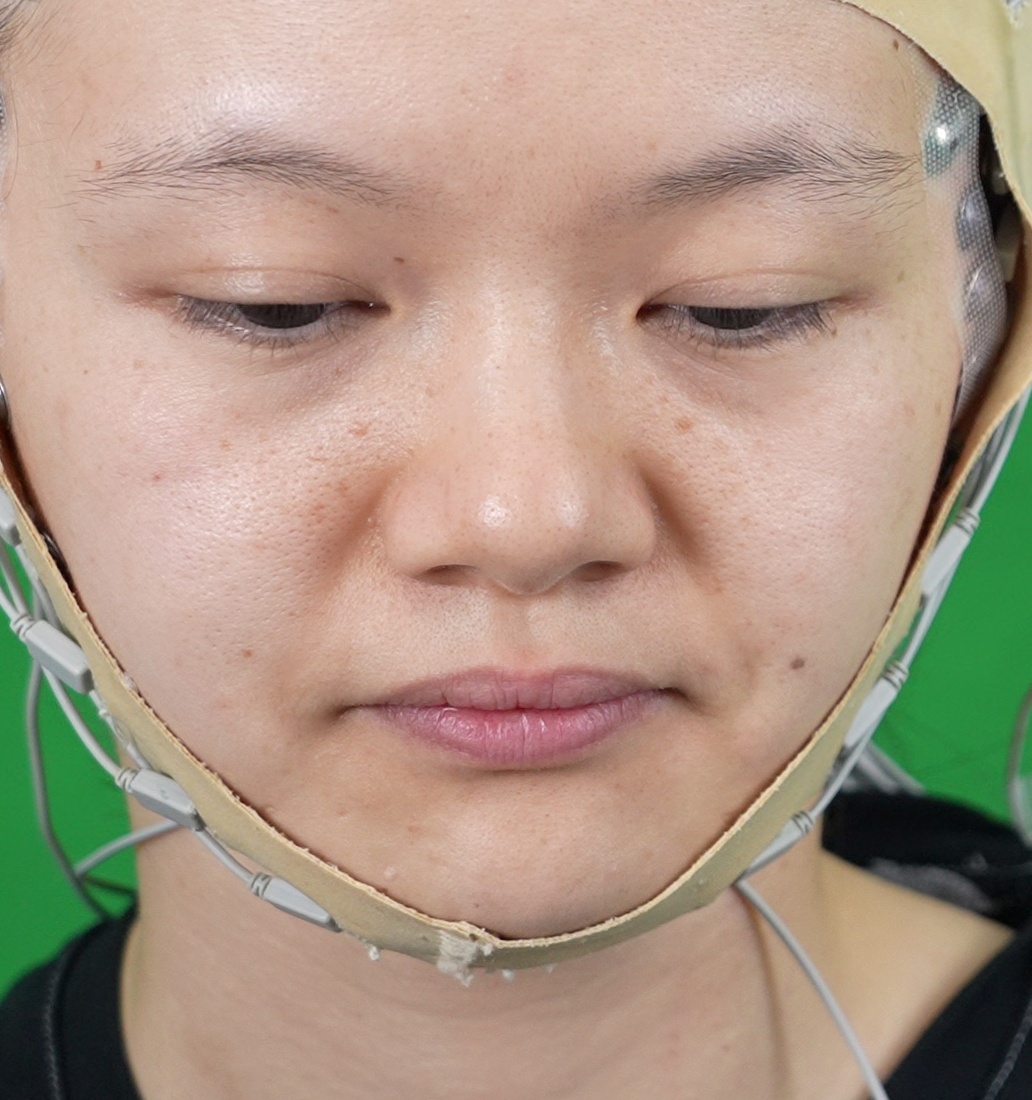}
    \label{fig:rgb-7}
  \end{subfigure}\hfill
  \begin{subfigure}{\mySubFigWidth}
    \includegraphics[width=\linewidth]{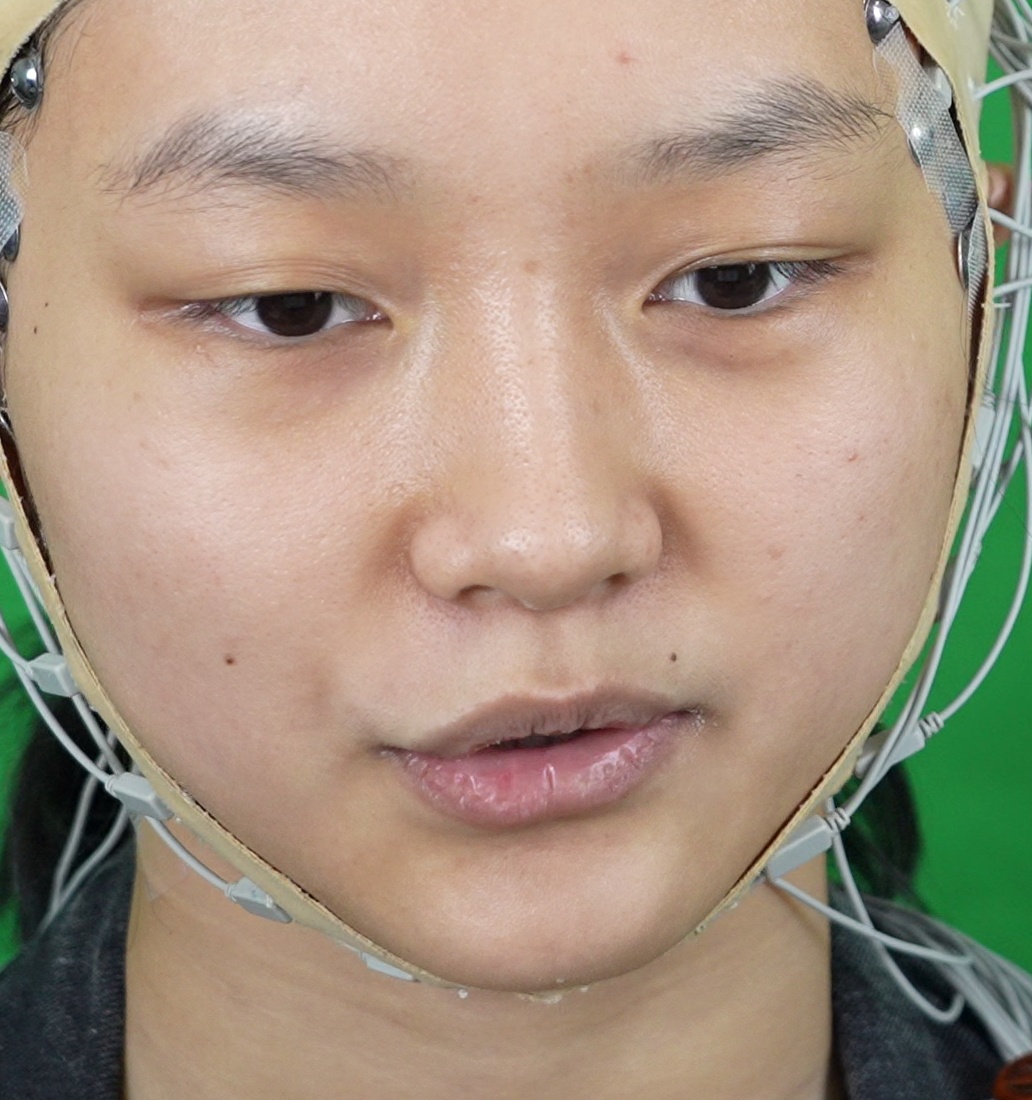}
    \label{fig:rgb-8}
  \end{subfigure}
  \vfill
  
  \begin{subfigure}{\mySubFigWidth}
    \includegraphics[width=\linewidth]{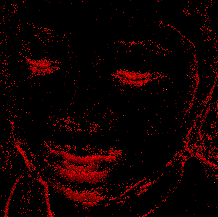}
    \label{fig:event-1}
  \end{subfigure}\hfill
  \begin{subfigure}{\mySubFigWidth}
    \includegraphics[width=\linewidth]{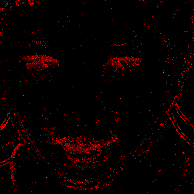}
    \label{fig:event-2}
  \end{subfigure}\hfill
  \begin{subfigure}{\mySubFigWidth}
    \includegraphics[width=\linewidth]{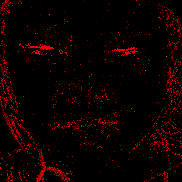}
    \label{fig:event-3}
  \end{subfigure}\hfill
  \begin{subfigure}{\mySubFigWidth}
    \includegraphics[width=\linewidth]{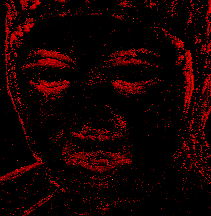}
    \label{fig:event-4}
  \end{subfigure}\hfill
  \begin{subfigure}{\mySubFigWidth}
    \includegraphics[width=\linewidth]{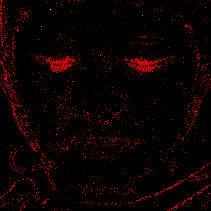}
    \label{fig:event-5}
  \end{subfigure}\hfill
  \begin{subfigure}{\mySubFigWidth}
    \includegraphics[width=\linewidth]{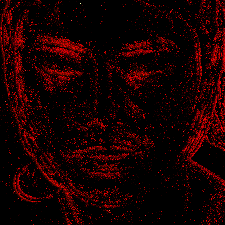}
    \label{fig:event-6}
  \end{subfigure}\hfill
  \begin{subfigure}{\mySubFigWidth}
    \includegraphics[width=\linewidth]{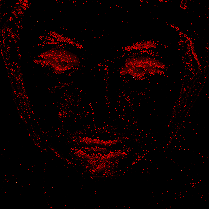}
    \label{fig:event-7}
  \end{subfigure}\hfill
  \begin{subfigure}{\mySubFigWidth}
    \includegraphics[width=\linewidth]{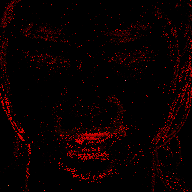}
    \label{fig:event-8}
  \end{subfigure}

  \caption{Spatial quality comparison. Top and bottom rows show RGB images and event-based frame, respectively.}
  \label{fig:spatial_quality_grid}
\end{figure}

\noindent\textbf{Quality Assessment.} We also perform a rigorous quality check on the 210 retaining high-quality aligned samples. Given the absence of external ground-truth for cross-modal synchronization in real-world setups, we evaluate alignment quality by quantifying the consistency between intensity frames and event streams.
\par
\noindent\textbf{1. Temporal Peak Correlation.} We hypothesize that ME dynamics exhibit synchronized peaks in both modalities near the apex. We construct a frame-difference-based RGB motion curve $m_R(t)$ and a bin-based event intensity curve $m_E(t)$. After z-score normalization and resampling to a unified length $L$, we perform a cross-correlation scan to find the optimal temporal lag $\delta^*$ and the maximum correlation coefficient $c^*$:
\begin{equation}
\delta^* = \arg \max_{l \in [-L_{max}, L_{max}]} \text{corr}(r_t, e_{t+l}), \quad c^* = \text{corr}(r_t, e_{t+\delta^*})
\end{equation}
Besides, we define the Peak Distance Ratio (PDR) as $PDR = |P_R - P_E|/ (L-1)$, where $P_R$ and $P_E$ denote the temporal indices of the respective curve peaks.
\par
\noindent\textbf{Quantitative Results.} On the 210 aligned samples, the median $c^*$ reaches 0.6222, while the median optimal physical lag $|\Delta t^*|$ is restricted to 0.0831 s. Notably, the 95th percentile of $c^*$ achieves 0.8848, indicating high temporal consistency across the majority of the dataset. The median PDR of 0.1471 further confirms that the most intense motion moments are well-synchronized. (Detailed per-sample metrics are provided in Supp.)
\par
\noindent\textbf{2. Spatial Alignment Validity.} Due to the fundamental differences in imaging mechanisms between RGB intensity and asynchronous event triggering, standard pixel-level metrics (e.g., SSIM, PSNR) are inherently low and unsuitable for cross-modal registration. Instead, we ensure spatial consistency by enforcing semantic ROI alignment through independent facial landmark detection and consistent cropping for both modalities. We provide qualitative ROI comparisons in \cref{fig:spatial_quality_grid} to demonstrate that both sensors focus on the same facial features.

\section{Benchmark Protocol and Results}
Due to the lack of publicly available source code for current event-based MER methods (e.g.,~\cite{guo2023gleffn,xiao2024estme}), a direct performance comparison is currently constrained. To address this, we establish a robust and reproducible baseline by integrating RGB and event feature extraction with our proposed CMC and CrossMamba modules. By releasing our complete pipeline, we aim to provide the community with the first standardized benchmark for cross-device event-RGB ME analysis.

\subsection{Baseline Method}
\begin{figure}[tb]
    \centering
    \includegraphics[width=0.75\linewidth]{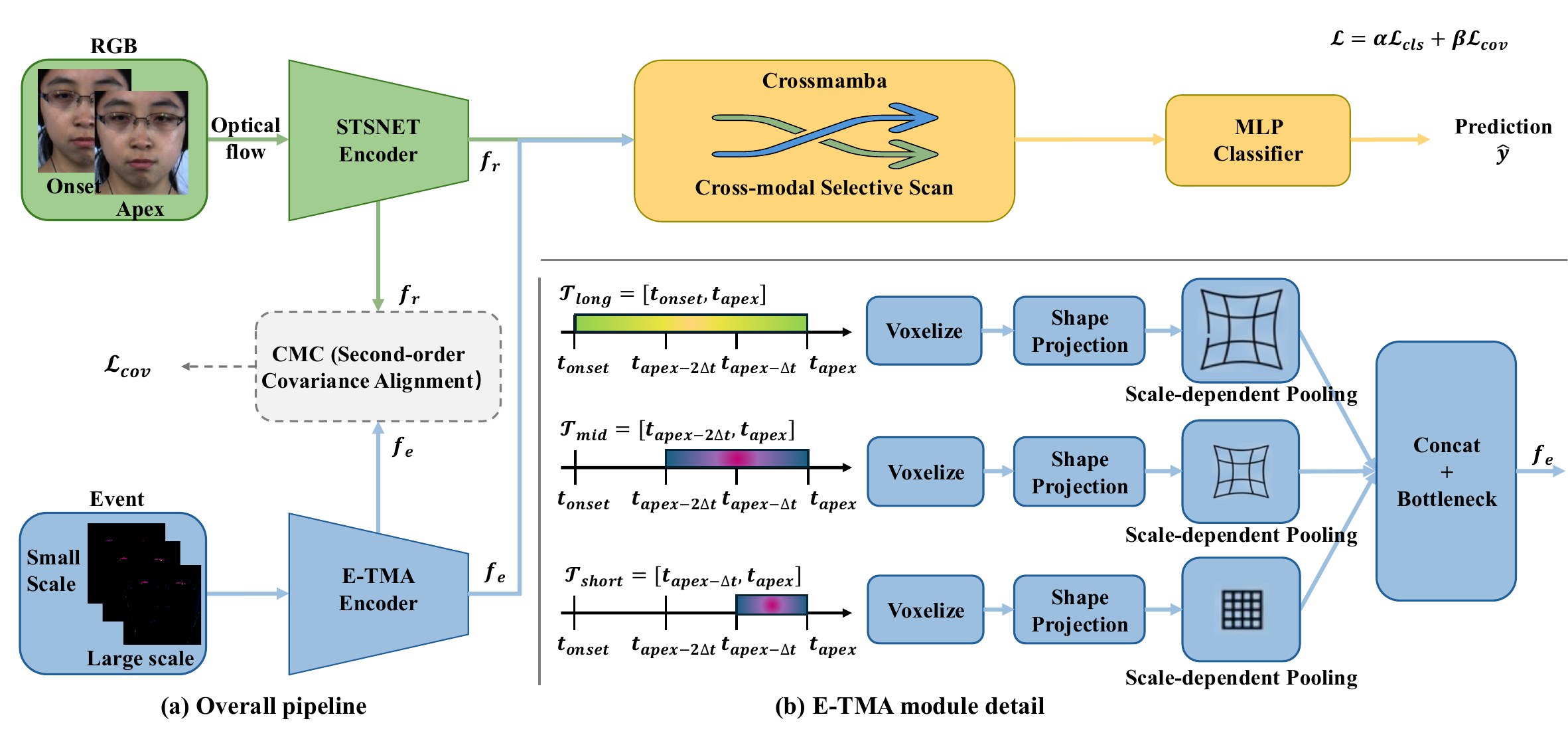}
    \caption{Overall architecture of the baseline method for multi-modal MER task}
    \label{fig:pipeline}
\end{figure}

As illustrated in \cref{fig:pipeline}, the proposed baseline integrates high-fidelity visual feature from RGB frames with high-temporal resolution asynchronous events. 

\noindent\textbf{A. Event-Temporal Multi-scale Aggregation (E-TMA).}
\label{sec:etma}

To bridge the sparsity-semantics gap in event streams, E-TMA aggregates evidence over asymmetric temporal windows anchored at the apex time $t_{apex}$.
\par
\noindent\textbf{Temporal Partitioning.} We partition $\mathcal{E}$ into three spans reflecting facial kinematics:
$\mathcal{T}_{long} = [t_{onset}, t_{apex}]$, $\mathcal{T}_{mid} = [t_{apex}-2\Delta t, t_{apex}]$,$ \mathcal{T}_{short} = [t_{apex}-\Delta t, t_{apex}]$, where $\Delta t$ is the RGB frame duration. These capture the complete trajectory, pre-peak acceleration, and peak impulsive signals, respectively.
\par
\noindent\textbf{Feature Aggregation.} Sub-streams $\mathcal{E}_{\mathcal{T}_i}$ are voxelized and mapped via shared projection $\Phi_{sh}(\cdot)$, followed by scale-dependent pooling $\mathcal{P}_{s_i}$ whose receptive field increases with window length to balance noise suppression and edge preservation:
$e'{i} = \mathcal{P}{s_i}\Big(\Phi_{sh}(\mathrm{Voxel}(\mathcal{E}{\mathcal{T}i}))\Big)$, $i \in {short, mid, long}$.
The final event representation $f_e$ is refined as:
$f_e = \mathrm{Bottleneck}\Big(\mathrm{Concat}(e'{s}, e'{m}, e'_{l})\Big)$.

\noindent\textbf{B. RGB Encoding and Cross-Modal Consistency (CMC).}

For RGB, we adopt a shallow STSTNet. Using optical flow $u, v$, and optical strain $\varepsilon$ between onset and apex , we extract $f_{rgb} = \text{Concat}(\text{Conv}_{3\times3\times3}(\phi))$ for $\phi \in \{u, v, \varepsilon\}$.
\par
To align the heterogeneous modalities, we introduce CMC, which couples the branches via second-order statistics to capture channel-wise co-activation patterns. For a batch of size $N$, we compute empirical covariance matrices $\Sigma^{\mathrm{RGB}}$ and $\Sigma^{\mathrm{Event}}$. The alignment objective minimizes the Frobenius distance is formed as: 
$\mathcal{L}{\mathrm{cov}}=\left|\Sigma^{\mathrm{RGB}}-\Sigma^{\mathrm{Event}}\right|{F}^{2}$.

\noindent\textbf{C. Bi-Directional CrossMamba Fusion and Optimization.}

Deep interaction is achieved via a Bi-Directional Cross-Selective State Space Model (Bi-CS-SSM). For a query $Q$ from one modality and keys/values $(K, V)$ from the other, the operation is defined as $
y_i = \sum_{j=1}^{i} Q_i K_j H_{i,j} V_j$,
where $H_{i,j}$ is the accumulated state transition. This allow RGB flow to filter event noise and event motion to localize RGB regions of interest. The total loss balances classification $\mathcal{L}_{cls}$ and structural alignment $\mathcal{L}_{cov}$, i.e.,
$\mathcal{L} = \alpha \mathcal{L}_{cls} + \beta \mathcal{L}_{cov}$, 
where $\alpha=0.8, \beta=0.2$ are determined by sensitivity analysis in Supp.

\subsection{Benchmark Result and Analysis}
In this subsection, we prioritize Benchmark with Tier-3, i.e., 210 aligned samples as our primary cross-modal benchmark due to its high alignment accuracy, which makes it particularly suitable for cross-modal research. Comparisons with single-modal SOTA methods are detailed in Supp. 
And we release Tier-2 (790 raw pairs) as (i) a challenging alignment testbed and (ii) a resource for learning event generation or domain adaptation from RGB under realistic sensor noise. We additionally report results on Benchmark with Tier-2 in the Suppl.

\noindent\textbf{Experimental Configuration.}
The aligned subset consists of 210 samples from 14 subjects (Happiness: 5\footnote{Regarding the class imbalance issue, our stress-induction paradigm is designed to preserve high ecological validity, and thus we do not enforce an artificial balance in the number of samples across emotion categories. This imbalance also reflects the natural distribution under stress-induction settings, where positive emotions occur less frequently. We will provide the confusion matrix in the Supp.}, Anger: 27, Disgust: 105, Surprise: 31 and Others: 42)\footnote{No samples satisfy the Fear rule in our current annotation set; thus Fear is merged into Class: Others. Additionally, the emotion distribution of the 720 samples is detailed in Supp.}. Following the standard protocol in ME research, we adopt the Leave-One-Subject-Out cross-validation protocol to ensure subject-independent evaluation. Performance is quantified using three standard metrics: Accuracy (ACC), Unweighted F1-score (UF1), and Unweighted Average Recall (UAR). Specific model parameters and configurations are in the Suppl.

\noindent\textbf{Main Results and Ablation Study.}
We conduct a series of ablation experiments to investigate the contribution of each modality and the proposed CMC module. The results are summarized in \cref{tab:ablation_cov}.
\begin{table}[tb]
\centering
\caption{Baseline performance and ablation study on the proposed aligned benchmark. }
\label{tab:ablation_cov}
\begin{tabular}{lccc}
\toprule
\textbf{Modality Setting} & \textbf{ACC} & \textbf{UF1} & \textbf{UAR} \\
\midrule
RGB only                 & 0.5762 & 0.3507 & 0.3443  \\
Event only               & 0.5619 & 0.3180 & 0.3167  \\
RGB + Event (w/o CMC)    & 0.5904 & 0.3628 & 0.3536  \\
\textbf{RGB + Event + CMC (Proposed baseline)} & \textbf{0.6667} & \textbf{0.4403} & \textbf{0.4246}  \\
\bottomrule
\end{tabular}
\end{table}

\par
\noindent\textit{Benchmark Validity.}
The RGB-only and event-only baselines achieve comparable performance, while simple fusion already improves UF1/UAR and the proposed consistency further boosts all metrics. This confirms that while RGB frames provide stable textural cues, the event modality captures transient onset-to-apex dynamics that are often missed or blurred in frame-based sampling, i.e., the two sensors offer complementary ME cues and our aligned pairs preserve meaningful cross-modal correspondence rather than introducing spurious gains.
\par
\noindent\textit{Efficacy of CMC.}
The integration of our CMC module yields a significant performance gain, boosting ACC to \textbf{0.6667} and UF1 to \textbf{0.4403}. This improvement indicates that simply concatenating features is insufficient for bridging the domain gap between synchronous frames and asynchronous events. By aligning the second-order correlation structures (covariance), CMC effectively stabilizes the shared representation against subject-specific variations and device-specific noise, leading to more robust emotion discriminative features.

\section{Conclusion}
In this paper, we present CDER-SME, a cross-device event-RGB micro-expression dataset collected under a structured multi-level stress induction framework. By simulating cognitive, and social stressors, our dataset bridges the gap between laboratory-controlled environments and high-stakes real-world scenarios. CDER-SME provides high-quality annotations including facial AUs and objective emotion labels derived from certified experts, which significantly promotes the development of affective intelligence in realistic contexts.
Another contribution of our work is the provision of a hardware-independent alignment pipeline that ensures temporal synchronization and semantic-level spatial registration for decoupled sensors. This setup reflects the practical challenges of deploying heterogeneous vision modules in unconstrained environments. To address the scarcity of open-source event-based MER methods, we establish a reproducible multimodal baseline based on the CrossMamba architecture, demonstrating the complementary benefits of high-frequency neuromorphic cues and rich textural details.
\par
Limitations and Future Work. While our rigorous synchronization criteria led to a reduction in the number of perfectly aligned bimodal pairs compared to the raw data pool, we release CDER-SME in three distinct tiers to maximize its utility for both single-modality training and high-precision cross-modal research. The current sample size constraints in aligned pairs slightly limit our ability to perform deep multi-task analysis on joint stress-expression prediction. In future work, we aim to develop more robust spatio-temporal registration algorithms to recover additional aligned samples from the raw streams. Furthermore, we will explore self-supervised pre-training on the full unaligned dataset to further enhance cross-modal micro-expression recognition in the wild.


%
%
\bibliographystyle{splncs04}
\bibliography{main}

@String(CVPR  = {IEEE Conf. Comput. Vis. Pattern Recog.})

@String(ICME  = {Int. Conf. Multimedia and Expo})

@String(CVPR  = {CVPR})

@String(ICME  =	{ICME})

@article{ekman1978facial,
  title={Facial action coding system},
  author={Ekman, Paul and Friesen, Wallace V},
  journal={Environmental Psychology \& Nonverbal Behavior},
  year={1978}
}

@book{picard1997affective,
  title={Affective computing},
  author={Picard, Rosalind W},
  year={1997},
  publisher={MIT press}
}

@inproceedings{pantic2005affective,
  title={Affective multimodal human-computer interaction},
  author={Pantic, Maja and Sebe, Nicu and Cohn, Jeffrey F and Huang, Thomas},
  booktitle={Proceedings of the 13th annual ACM international conference on Multimedia},
  pages={669--676},
  year={2005}
}

@book{ekman2009telling,
  title={Telling lies: Clues to deceit in the marketplace politics and marriage},
  author={Ekman, Paul},
  year={2009},
  publisher={WW Norton \& Company}
}

@article{li2025could,
  author={Li, Jingting and Lu, Shaoyuan and Wang, Yan and Dong, Zizhao and Wang, Su-Jing and Fu, Xiaolan},
  journal={IEEE Transactions on Affective Computing}, 
  title={Could Micro-Expressions Be Quantified? Electromyography Gives Affirmative Evidence}, 
  year={2025},
  volume={16},
  number={4},
  pages={2959-2974},
  doi={10.1109/TAFFC.2025.3575127}}

@article{ekman2009lie,
  title={Lie catching and microexpressions},
  author={Ekman, Paul},
  journal={The philosophy of deception},
  volume={1},
  number={2},
  pages={5},
  year={2009}
}

@article{porter2008reading,
  title={Reading between the lies: Identifying concealed and falsified emotions in universal facial expressions},
  author={Porter, Stephen and Ten Brinke, Leanne},
  journal={Psychological science},
  volume={19},
  number={5},
  pages={508--514},
  year={2008},
  publisher={SAGE Publications Sage CA: Los Angeles, CA}
}

@article{li2022deep,
  title={Deep learning for micro-expression recognition: A survey},
  author={Li, Yante and Wei, Jinsheng and Liu, Yang and Kauttonen, Janne and Zhao, Guoying},
  journal={IEEE Transactions on Affective Computing},
  volume={13},
  number={4},
  pages={2028--2046},
  year={2022},
  publisher={IEEE}
}

@article{yan2014casme2,
	title = {{CASME} {II}: An Improved Spontaneous Micro-Expression Database and the Baseline Evaluation},
	volume = {9},
	issn = {1932-6203},
	url = {https://dx.plos.org/10.1371/journal.pone.0086041},
	doi = {10.1371/journal.pone.0086041},
	shorttitle = {{CASME} {II}},
	pages = {e86041},
	number = {1},
	journal = {{PLoS} {ONE}},
	shortjournal = {{PLoS} {ONE}},
	author = {Yan, Wen-Jing and Li, Xiaobai and Wang, Su-Jing and Zhao, Guoying and Liu, Yong-Jin and Chen, Yu-Hsin and Fu, Xiaolan},
	editor = {Guo, Kun},
	urldate = {2024-08-02},
	year = {2014-01-27},
}

@INPROCEEDINGS{li2013smic,
  author={Li, Xiaobai and Pfister, Tomas and Huang, Xiaohua and Zhao, Guoying and Pietikäinen, Matti},
  booktitle={2013 10th IEEE International Conference and Workshops on Automatic Face and Gesture Recognition (FG)}, 
  title={{A Spontaneous Micro-expression Database: Inducement, collection and baseline}}, 
  year={2013},
  volume={},
  number={},
  pages={1-6},
  doi={10.1109/FG.2013.6553717}}

@inproceedings{yan2013casme,
	location = {Shanghai, China},
	title = {{CASME} database: A dataset of spontaneous micro-expressions collected from neutralized faces},
	isbn = {978-1-4673-5546-9 978-1-4673-5545-2 978-1-4673-5544-5},
	url = {http://ieeexplore.ieee.org/document/6553799/},
	doi = {10.1109/FG.2013.6553799},
	shorttitle = {{CASME} database},
	eventtitle = {2013 10th {IEEE} International Conference on Automatic Face \& Gesture Recognition ({FG} 2013)},
	pages = {1--7},
	booktitle = {2013 10th {IEEE} International Conference and Workshops on Automatic Face and Gesture Recognition ({FG})},
	publisher = {{IEEE}},
	author = {{Wen-Jing Yan} and Wu, Qi and {Yong-Jin Liu} and {Su-Jing Wang} and Fu, Xiaolan},
	urldate = {2024-08-02},
	year = {2013},
}

@article{li2022cas3,
  title={{CAS(ME)$^3$: A Third Generation Facial Spontaneous Micro-Expression Database With Depth Information and High Ecological Validity}},
  author={Li, Jingting and Dong, Zizhao and Lu, Shaoyuan and Wang, Su-Jing and Yan, Wen-Jing and Ma, Yinhuan and Liu, Ye and Huang, Changbing and Fu, Xiaolan},
  journal={IEEE Transactions on Pattern Analysis and Machine Intelligence},
  volume={45},
  number={3},
  pages={2782--2800},
  year={2022},
  publisher={IEEE}
}

@article{Davison2018samm,
	title = {{SAMM}: {A} {S}pontaneous {M}icro-{F}acial {M}ovement {D}ataset},
	volume = {9},
	issn = {1949-3045},
	url = {https://ieeexplore.ieee.org/document/7492264/?arnumber=7492264},
	doi = {10.1109/TAFFC.2016.2573832},
	shorttitle = {{SAMM}},
	pages = {116--129},
	number = {1},
	journal = {{IEEE} Transactions on Affective Computing},
	author = {Davison, Adrian K. and Lansley, Cliff and Costen, Nicholas and Tan, Kevin and Yap, Moi Hoon},
	urldate = {2024-08-02},
	year = {2018},
}

@ARTICLE{ben2021video,
  author={Ben, Xianye and Ren, Yi and Zhang, Junping and Wang, Su-Jing and Kpalma, Kidiyo and Meng, Weixiao and Liu, Yong-Jin},
  journal={IEEE Transactions on Pattern Analysis and Machine Intelligence}, 
  title={{V}ideo-{B}ased {F}acial {M}icro-{E}xpression {A}nalysis: {A} {S}urvey of {D}atasets, {F}eatures and {A}lgorithms}, 
  year={2022},
  volume={44},
  number={9},
  pages={5826-5846},
  doi={10.1109/TPAMI.2021.3067464}}

@article{li20234DME,
	title = {{4DME: A Spontaneous 4D Micro-Expression Dataset With Multimodalities}},
	volume = {14},
	issn = {1949-3045},
	url = {https://ieeexplore.ieee.org/document/9796028/?arnumber=9796028},
	doi = {10.1109/TAFFC.2022.3182342},
	shorttitle = {4DME},
	pages = {3031--3047},
	number = {4},
	journal = {{IEEE} Transactions on Affective Computing},
	author = {Li, Xiaobai and Cheng, Shiyang and Li, Yante and Behzad, Muzammil and Shen, Jie and Zafeiriou, Stefanos and Pantic, Maja and Zhao, Guoying},
	urldate = {2024-08-02},
	year = {2023},
}

@article{zhao2024DFME,
	title = {{DFME}: {A} {N}ew {B}enchmark for {D}ynamic {F}acial {M}icro-{E}xpression {R}ecognition},
	journal = {{IEEE} Transactions on Affective Computing},
	shortjournal = {{IEEE} Trans. Affective Comput.},
	author = {Zhao, Sirui and Tang, Huaying and Mao, Xinglong and Liu, Shifeng and Zhang, Yiming and Wang, Hao and Xu, Tong and Chen, Enhong},
	year={2024},
  volume={15},
  number={3},
  pages={1371-1386},
  doi={10.1109/TAFFC.2023.3341918}}

@inproceedings{husak2017spotting,
  title={Spotting facial micro-expressions “in the wild”},
  author={Hus{\'a}k, Petr and Cech, Jan and Matas, Ji{\v{r}}{\'\i}},
  booktitle={22nd Computer Vision Winter Workshop (Retz)},
  pages={1--9},
  year={2017}
}

@article{dickerson2004acute,
  title={Acute stressors and cortisol responses: a theoretical integration and synthesis of laboratory research},
  author={Dickerson, Sally S and Kemeny, Margaret E},
  journal={Psychological bulletin},
  volume={130},
  number={3},
  pages={355},
  year={2004},
  publisher={American Psychological Association}
}

@article{kirschbaum1993trier,
  title={The ‘Trier Social Stress Test’--a tool for investigating psychobiological stress responses in a laboratory setting},
  author={Kirschbaum, Clemens and Pirke, Karl-Martin and Hellhammer, Dirk H},
  journal={Neuropsychobiology},
  volume={28},
  number={1-2},
  pages={76--81},
  year={1993},
  publisher={S. Karger AG Basel, Switzerland}
}

@article{gallego2020event,
  title={Event-based vision: A survey},
  author={Gallego, Guillermo and Delbr{\"u}ck, Tobi and Orchard, Garrick and Bartolozzi, Chiara and Taba, Brian and Censi, Andrea and Leutenegger, Stefan and Davison, Andrew J and Conradt, J{\"o}rg and Daniilidis, Kostas and others},
  journal={IEEE transactions on pattern analysis and machine intelligence},
  volume={44},
  number={1},
  pages={154--180},
  year={2020},
  publisher={IEEE}
}

@inproceedings{berlincioni2023neuromorphic,
  title={Neuromorphic Event-based Facial Expression Recognition},
  author={Berlincioni, L. and others},
  booktitle={Proceedings of the IEEE/CVF Conference on Computer Vision and Pattern Recognition (CVPR) Workshops},
  pages={4109--4119},
  year={2023}
}

@inproceedings{guo2023gleffn,
  title={{GLEFFN}: A Global-Local Event Feature Fusion Network for Micro-Expression Recognition},
  author={Guo, Cunhan and Huang, Heyan},
  booktitle={Proceedings of the 3rd Workshop on Facial Micro-Expression (FME ’23)},
  pages={8},
  year={2023},
  publisher={ACM}
}

@inproceedings{xiao2024estme,
  title={{ESTME}: Event-driven Spatio-temporal Motion Enhancement for Micro-Expression Recognition},
  author={Xiao, P. and others},
  booktitle={IEEE International Conference on Multimedia and Expo (ICME)},
  pages={1--6},
  year={2024}
}

@inproceedings{mastropasqua2025exploring,
  title={Exploring spatial-temporal dynamics in event-based facial micro-expression analysis},
  author={Mastropasqua, Nicolas and Bugueno-Cordova, Ignacio and Verschae, Rodrigo and Acevedo, Daniel and Negri, Pablo and Buemi, Maria Elena},
  booktitle={Proceedings of the IEEE/CVF International Conference on Computer Vision},
  pages={4723--4732},
  year={2025}
}

@inproceedings{becattini2024neuromorphic,
  title={Neuromorphic facial analysis with cross-modal supervision},
  author={Becattini, Federico and Cultrera, Luca and Berlincioni, Lorenzo and Ferrari, Claudio and Leonardo, Andrea and Del Bimbo, Alberto},
  booktitle={European Conference on Computer Vision},
  pages={205--223},
  year={2024},
  organization={Springer}
}

@inproceedings{adra2024beyond,
  title={Beyond {RGB}: Tri-modal microexpression recognition with rgb, thermal, and event data},
  author={Adra, Mira and Mirabet-Herranz, Nelida and Dugelay, Jean-Luc},
  booktitle={International Conference on Pattern Recognition},
  pages={311--324},
  year={2024},
  organization={Springer}
}

@inproceedings{shreve2011macro,
  title={Macro-and micro-expression spotting in long videos using spatio-temporal strain},
  author={Shreve, Matthew and Godavarthy, Sridhar and Goldgof, Dmitry and Sarkar, Sudeep},
  booktitle={2011 IEEE international conference on automatic face \& gesture recognition (FG)},
  pages={51--56},
  year={2011},
  organization={IEEE}
}

@inproceedings{polikovsky2009facial,
  title={Facial micro-expressions recognition using high speed camera and 3D-gradient descriptor},
  author={Polikovsky, Senya and Kameda, Yoshinari and Ohta, Yuichi},
  booktitle={3rd international conference on imaging for crime detection and prevention (ICDP 2009)},
  pages={1--6},
  year={2009},
  organization={IET}
}

@article{stroop1935studies,
  title={Studies of interference in serial verbal reactions.},
  author={Stroop, J Ridley},
  journal={Journal of experimental psychology},
  volume={18},
  number={6},
  pages={643},
  year={1935},
  publisher={Psychological Review Company}
}

@article{dong2022spontaneous,
  title={Spontaneous facial expressions and micro-expressions coding: from brain to face},
  author={Dong, Zizhao and Wang, Gang and Lu, Shaoyuan and Li, Jingting and Yan, Wenjing and Wang, Su-Jing},
  journal={Frontiers in Psychology},
  volume={12},
  pages={784834},
  year={2022},
  publisher={Frontiers Media SA}
}

@article{davison2018objective,
  title={Objective classes for micro-facial expression recognition},
  author={Davison, Adrian K and Merghani, Walied and Yap, Moi Hoon},
  journal={Journal of imaging},
  volume={4},
  number={10},
  pages={119},
  year={2018},
  publisher={MDPI}
}

@article{he2022micro,
  title={Micro-expression spotting based on optical flow features},
  author={He, Yuhong and Xu, Zhongliang and Ma, Lin and Li, Haifeng},
  journal={Pattern Recognition Letters},
  volume={163},
  pages={57--64},
  year={2022},
  publisher={Elsevier}
}

@article{rebecq2019high,
  title={High speed and high dynamic range video with an event camera},
  author={Rebecq, Henri and Ranftl, Ren{\'e} and Koltun, Vladlen and Scaramuzza, Davide},
  journal={IEEE transactions on pattern analysis and machine intelligence},
  volume={43},
  number={6},
  pages={1964--1980},
  year={2019},
  publisher={IEEE}
}

@inproceedings{lin2022dvs,
  title={Dvs-voltmeter: Stochastic process-based event simulator for dynamic vision sensors},
  author={Lin, Songnan and Ma, Ye and Guo, Zhenhua and Wen, Bihan},
  booktitle={European Conference on Computer Vision},
  pages={578--593},
  year={2022},
  organization={Springer}
}

@article{li2022mmnet,
  title={MMNet: Muscle motion-guided network for micro-expression recognition},
  author={Li, Hanting and Sui, Mingzhe and Zhu, Zhaoqing and Zhao, Feng},
  journal={arXiv preprint arXiv:2201.05297},
  year={2022}
}

@article{chen2022block,
  title={Block division convolutional network with implicit deep features augmentation for micro-expression recognition},
  author={Chen, Bin and Liu, Kun-Hong and Xu, Yong and Wu, Qing-Qiang and Yao, Jun-Feng},
  journal={IEEE Transactions on Multimedia},
  volume={25},
  pages={1345--1358},
  year={2022},
  publisher={IEEE}
}

@article{wang2024htnet,
  title={Htnet for micro-expression recognition},
  author={Wang, Zhifeng and Zhang, Kaihao and Luo, Wenhan and Sankaranarayana, Ramesh},
  journal={Neurocomputing},
  volume={602},
  pages={128196},
  year={2024},
  publisher={Elsevier}
}

\newpage
\section*{Supplementary Materials}
\appendix
\section{Dataset Release Protocol and Data Card}

To facilitate community use and reproducibility, we organize CDER-SME into three release tiers with different quality-control levels and intended use cases. The dataset is designed not only as a collection of samples, but also as a structured benchmark package with annotations, metadata, and official evaluation utilities. CDER-SME contains 1,963 RGB samples from 92 subjects in total, including 790 raw Event–RGB pairs and 210 high-fidelity aligned Event–RGB pairs. The three-tier organization follows the benchmark design described in the main paper.

Tier-1 contains all 1,963 RGB samples and is intended for single-modality RGB-based training, pretraining, and stress-related facial behavior analysis. Tier-2 contains the 790 raw Event–RGB pairs recorded under the decoupled cross-device setting. This tier preserves realistic synchronization noise, marker instability, and acquisition imperfections, and is therefore suitable for research on weak synchronization, cross-modal adaptation, event generation from RGB, and robust alignment. Tier-3 contains the 210 high-precision aligned Event–RGB pairs that passed strict quality-control filtering and forms the official benchmark core for high-fidelity cross-modal micro-expression recognition (MER). The filtering process and exclusion accounting are detailed in the supplementary statistics.

For each released sample, we provide the original modality data together with structured annotations and benchmark metadata. Specifically, the release package includes:
\begin{enumerate}
    \item raw RGB videos and raw event streams in their original formats;
     \item temporal annotations including onset, apex, and offset;
  \item AU annotations and AU laterality labels;
 \item derived objective emotion labels based on AU rules;
  \item stress-condition labels indicating the corresponding induction setting;
 \item alignment metadata including estimated cross-device temporal offset, face-crop information;
  \item official split files, evaluation scripts, and reference-baseline code. This release design follows the benchmark plan stated in the main paper and aims to support transparent and reproducible downstream comparisons. 
\end{enumerate}

In practice, we recommend Tier-1 for large-scale RGB-only learning, Tier-2 for robust cross-device or weakly synchronized multimodal research, and Tier-3 for standardized benchmark evaluation. This tiered release strategy is especially important because the decoupled acquisition setup reflects realistic deployment conditions, while the aligned subset provides a high-confidence core for fair comparison across multimodal methods.

\section{Ethics, Privacy, and Responsible Release}

CDER-SME contains human facial recordings collected in stress-induction settings, and we therefore treat ethics, privacy, and responsible release as an integral part of the benchmark design. As stated in the main paper, all experimental protocols and data collection procedures were approved by the Institutional Review Board, and written informed consent was obtained from all 92 participants prior to data collection. Participants explicitly granted permission for their facial data to be used for scientific research.

Because the dataset involves identifiable facial information and stress-related behavioral responses, the released benchmark package will focus on research utility while limiting unnecessary exposure of personal information. We do not associate samples with real names or other direct identity metadata in the released annotations. The provided metadata are restricted to research-relevant fields such as modality data, temporal labels, AU labels, emotion labels, stress-condition labels, and alignment information. This design aims to support reproducible scientific use while minimizing privacy risks. The benchmark package also emphasizes objective coding based on observed facial actions rather than self-disclosed personal information.

The social-stress module includes deceptive-speech conditions intended to induce emotional leakage under evaluative pressure. For this reason, the dataset should be interpreted as a benchmark for affective computing and cross-modal ME analysis, rather than as a tool for real-world lie detection or any high-stakes automatic decision-making. We release the dataset to advance research on subtle facial dynamics, multimodal alignment, and neuromorphic affect sensing under controlled scientific conditions. Consistent with this goal, the benchmark design separates a high-fidelity aligned subset from a noisier raw-pair subset and provides transparent documentation of filtering and annotation procedures.

We believe that a responsible dataset release must combine ethical approval, informed consent, limited metadata exposure, and transparent benchmark protocols. CDER-SME is released with this principle in mind, and the supplementary material is intended to document these safeguards alongside the scientific contribution of the dataset.

\section{Data Collection Setup Illustration}

\begin{figure}[tb]
    \centering
    \includegraphics[width=0.9\linewidth]{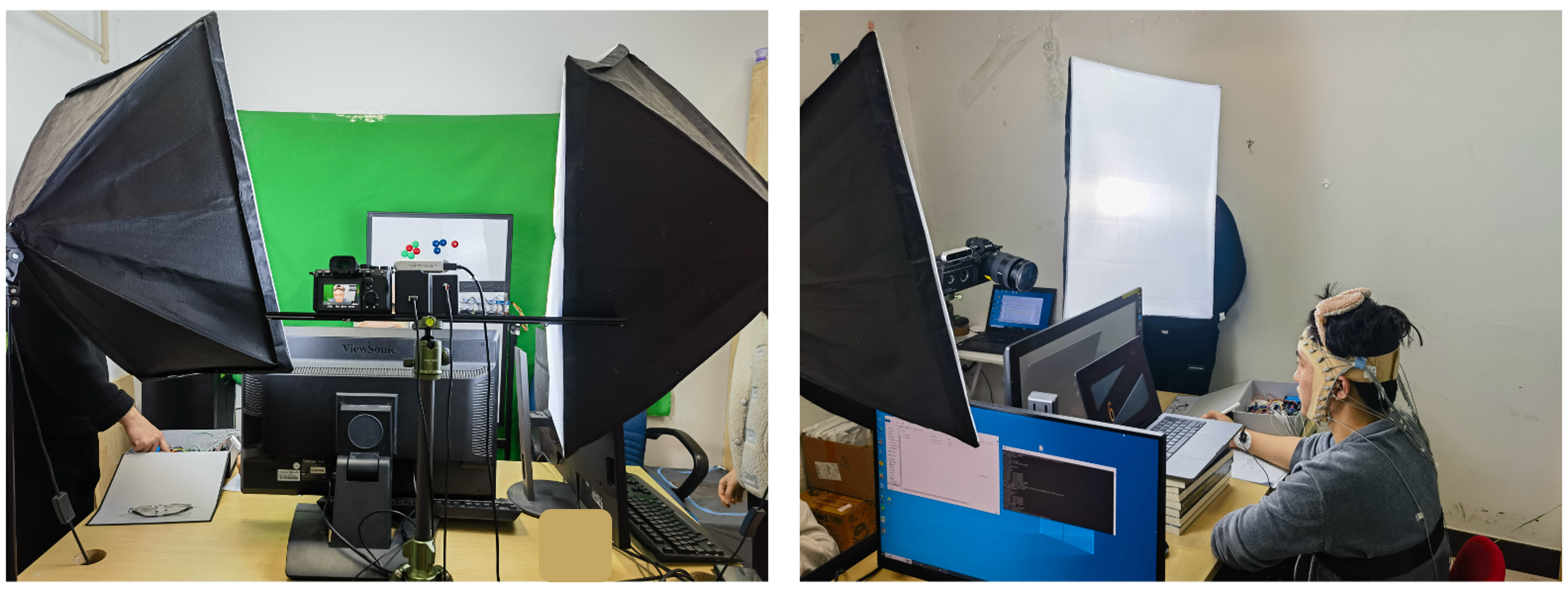}
    \caption{Data collection setup}
    \label{fig:setup}
\end{figure}

Our data were collected in an indoor laboratory setting with a controlled, repeatable recording rig. As shown in \cref{fig:setup}, participants were seated at a fixed distance in front of a stimulus monitor, while the acquisition devices were mounted on a rigid support/tripod and aligned to the subject’s frontal face at approximately eye level. To reduce illumination variance and motion-induced artifacts, we used a two-point soft-lighting setup with large softboxes/diffusion panels on both sides of the subject, producing uniform, shadow-minimized facial illumination. A green-screen/background panel was placed behind the monitor to provide a clean, consistent backdrop and to limit background clutter in the field of view. 
All recordings were performed at a dedicated workstation with real-time monitoring to ensure stable framing and recording quality across devices throughout the stress-induction tasks. Overall, this setup was designed not merely to illustrate the recording environment, but to support controlled-yet-realistic acquisition with reduced illumination variation, suppressed event noise, and reproducible recording conditions across subjects and sessions. Importantly, consistent with the main paper, the RGB and event sensors were deployed as independent, decoupled devices rather than as a rigidly coaxial or pre-calibrated hardware pair, so that the resulting dataset better reflects practical cross-device Event–RGB acquisition in realistic deployment settings.

\section{Psychological Experiment Statistic Analysis}

\subsection{Analysis of the Cognitive-Stress Task}
\label{sec:stroop_stress}

To investigate whether facial expression behaviors varied across different levels of cognitive stress, we first conducted chi-square tests on the numbers of micro-expressions (MEs) and macro-expressions (MaEs). The results showed that the number of MEs did not differ significantly across the three stress levels ($\chi^2(1,28)=29.351$, $p=0.395$). Pairwise comparisons also showed no significant differences between the no-stress and low-stress conditions ($\chi^2(1,42)=46.751$, $p=0.284$), the high-stress and low-stress conditions ($\chi^2(1,70)=52.771$, $p=0.938$), or the no-stress and high-stress conditions ($\chi^2(1,60)=48.919$, $p=0.846$). Similarly, the number of MaEs did not differ significantly across the three stress levels ($\chi^2(1,26)=31.909$, $p=0.196$). Pairwise comparisons likewise revealed no significant differences between no stress and low stress ($\chi^2(1,56)=73.237$, $p=0.061$), high stress and low stress ($\chi^2(1,63)=64.988$, $p=0.408$), or no stress and high stress ($\chi^2(1,72)=70.755$, $p=0.519$).
\par
We then examined the facial action units (AUs) that appeared more frequently under high cognitive stress. According to the annotation statistics (\cref{tab:stroop_au_ranking}), five AUs occurred more frequently in the high-stress condition, suggesting that facial movements under cognitive stress were concentrated more strongly in the upper face, especially around the orbicularis oculi region, such as AU5 (upper lid raiser) and AU7 (lid tightener). This observation is consistent with prior findings suggesting that upper-face muscles may leak emotional states in a relatively involuntary manner.
\begin{table}[tb]
\centering
\caption{Frequency rankings of facial action units in the Cognitive Stress Task - high-stress Condition}
\label{tab:stroop_au_ranking}
\begin{tabular}{cccc}
\toprule
\textbf{Rank} & \textbf{All AU} & \textbf{MaE AU} & \textbf{ME AU} \\
\midrule
1 & AU5  & AU5  & AU7  \\
2 & AU7  & AU24 & AU4  \\
3 & AU4  & AU2  & AU5  \\
4 & AU24 & AU7  & AU24 \\
5 & AU2  & AU4  & AU2  \\
\bottomrule
\end{tabular}
\end{table}
\par
To further test whether AU counts varied across the three cognitive stress levels, we conducted one-way ANOVA (\cref{tab:stroop_au_combined}). The total number of AUs differed significantly across stress levels ($F(2,88)=4.596$, $p<0.05$, $\eta_p^2=0.095$). Among the AUs that appeared frequently under high cognitive stress, AU4 (brow lowerer) showed a significant difference across the three levels ($F(2,88)=6.340$, $p<0.01$, $\eta_p^2=0.126$), and AU7 (lid tightener) also differed significantly ($F(2,88)=13.543$, $p<0.001$, $\eta_p^2=0.235$). In contrast, AU2 (outer brow raiser; $F(2,88)=2.190$, $p=0.118$, $\eta_p^2=0.047$), AU5 (upper lid raiser; $F(2,88)=2.870$, $p=0.062$, $\eta_p^2=0.061$), and AU24 (lip pressor; $F(2,88)=1.101$, $p=0.337$, $\eta_p^2=0.024$) did not show significant differences across stress levels.

\begin{table}[tb]
\centering
\caption{Facial AU statistics in the Stroop-based Cognitive Stress Task across three stress levels. HS, LS, and NS denote High Stress, Low Stress, and No Stress, respectively.}
\label{tab:stroop_au_combined}
\setlength{\tabcolsep}{5pt}
\begin{tabular}{lccc ccc cc}
\toprule
& \multicolumn{3}{c}{\textbf{Count}} & \multicolumn{3}{c}{\textbf{Mean $\pm$ Std}} & \multirow{2}{*}{\textbf{F}} & \multirow{2}{*}{\boldmath$\eta_p^2$} \\
\cmidrule(lr){2-4} \cmidrule(lr){5-7}
\textbf{AU} & \textbf{HS} & \textbf{LS} & \textbf{NS} & \textbf{HS} & \textbf{LS} & \textbf{NS} &  &  \\
\midrule
AU1  & 27 & 13 & 27 & 0.96$\pm$0.96 & 0.42$\pm$0.77 & 0.84$\pm$1.71 & 1.637     & 0.036 \\
AU2  & 40 & 20 & 26 & 1.43$\pm$1.83 & 0.65$\pm$0.99 & 0.81$\pm$1.60 & 2.190     & 0.047 \\
AU4  & 59 & 17 & 21 & 2.11$\pm$2.81 & 0.55$\pm$1.09 & 0.66$\pm$1.36 & 6.340**   & 0.126 \\
AU5  & 61 & 29 & 36 & 2.18$\pm$3.08 & 0.94$\pm$1.06 & 1.13$\pm$1.88 & 2.870     & 0.061 \\
AU6  & 2  & 0  & 0  & 0.07$\pm$0.26 & 0.00$\pm$0.00 & 0.00$\pm$0.00 & 2.343     & 0.051 \\
AU7  & 65 & 24 & 12 & 2.32$\pm$2.26 & 0.77$\pm$1.28 & 0.38$\pm$0.66 & 13.543*** & 0.235 \\
AU9  & 3  & 8  & 6  & 0.11$\pm$0.42 & 0.26$\pm$0.63 & 0.19$\pm$0.64 & 0.500     & 0.011 \\
AU10 & 9  & 2  & 3  & 0.32$\pm$1.19 & 0.06$\pm$0.25 & 0.09$\pm$0.53 & 1.038     & 0.023 \\
AU11 & 0  & 4  & 0  & 0.00$\pm$0.00 & 0.13$\pm$0.56 & 0.00$\pm$0.00 & 1.579     & 0.035 \\
AU12 & 8  & 2  & 1  & 0.29$\pm$0.71 & 0.06$\pm$0.25 & 0.03$\pm$0.18 & 2.967     & 0.063 \\
AU14 & 31 & 19 & 31 & 1.11$\pm$1.47 & 0.61$\pm$1.26 & 0.97$\pm$1.18 & 1.151     & 0.025 \\
AU15 & 1  & 2  & 0  & 0.04$\pm$0.19 & 0.06$\pm$0.36 & 0.00$\pm$0.00 & 0.599     & 0.013 \\
AU17 & 20 & 22 & 23 & 0.71$\pm$0.94 & 0.71$\pm$0.97 & 0.72$\pm$1.55 & 0.000     & 0.000 \\
AU20 & 34 & 20 & 20 & 1.21$\pm$1.87 & 0.65$\pm$1.25 & 0.63$\pm$1.72 & 1.228     & 0.027 \\
AU22 & 3  & 0  & 0  & 0.11$\pm$0.42 & 0.00$\pm$0.00 & 0.00$\pm$0.00 & 2.093     & 0.045 \\
AU23 & 5  & 10 & 6  & 0.18$\pm$0.48 & 0.32$\pm$0.95 & 0.19$\pm$0.78 & 0.338     & 0.008 \\
AU24 & 55 & 39 & 40 & 1.96$\pm$2.22 & 1.26$\pm$1.37 & 1.25$\pm$2.55 & 1.101     & 0.024 \\
AU25 & 9  & 2  & 1  & 0.32$\pm$1.34 & 0.06$\pm$0.25 & 0.03$\pm$0.18 & 1.270     & 0.028 \\
AU26 & 13 & 11 & 12 & 0.46$\pm$0.84 & 0.35$\pm$1.11 & 0.38$\pm$0.98 & 0.101     & 0.002 \\
AU28 & 3  & 4  & 6  & 0.11$\pm$0.42 & 0.13$\pm$0.34 & 0.19$\pm$0.47 & 0.308     & 0.007 \\
AU38 & 2  & 0  & 0  & 0.07$\pm$0.26 & 0.00$\pm$0.00 & 0.00$\pm$0.00 & 2.343     & 0.051 \\
AU41 & 1  & 4  & 0  & 0.04$\pm$0.19 & 0.13$\pm$0.72 & 0.00$\pm$0.00 & 0.741     & 0.017 \\
AU43 & 11 & 4  & 6  & 0.39$\pm$0.57 & 0.13$\pm$0.56 & 0.19$\pm$0.47 & 1.962     & 0.043 \\
AU44 & 5  & 3  & 5  & 0.18$\pm$0.48 & 0.10$\pm$0.40 & 0.16$\pm$0.45 & 0.277     & 0.006 \\
\midrule
Total & -- & -- & -- & 16.68$\pm$12.49 & 8.35$\pm$7.95 & 8.81$\pm$13.99 & 4.596* & 0.095 \\
\bottomrule
\end{tabular}
\vspace{2mm}
\begin{minipage}{0.98\linewidth}
\footnotesize
\textit{Note:} * $p<0.05$, ** $p<0.01$, *** $p<0.001$.
\end{minipage}
\end{table}

\subsection{Analysis of the Social-Stress Task}
\label{sec:active_deception}

We next analyzed facial expressions under social-stress conditions induced by \emph{reading}, \emph{speech}, and \emph{lying}. Chi-square tests showed that the overall number of MEs did not differ significantly across the three conditions ($\chi^2(1,40)=49.619$, $p=0.142$). However, pairwise comparisons revealed significant differences between reading and speech ($\chi^2(1,165)=273.378$, $p<0.001$), between lying and speech ($\chi^2(1,270)=375.222$, $p<0.001$), and between reading and lying ($\chi^2(1,198)=263.823$, $p<0.01$). For MaEs, the overall difference across the three conditions was significant ($\chi^2(1,28)=72.837$, $p<0.001$). Pairwise comparisons were also significant between reading and speech ($\chi^2(1,55)=96.960$, $p<0.001$), between lying and speech ($\chi^2(1,143)=175.426$, $p<0.05$), and between reading and lying ($\chi^2(1,60)=106.684$, $p<0.001$).
\par
Based on the AU annotations (\cref{tab:social_pressor_au_ranking}), several AUs occurred more frequently under high social stress. These AUs were mainly associated with the frontalis muscle region and the orbicularis oris region, including AU1 (inner brow raiser), AU2 (outer brow raiser), and AU24 (lip pressor), suggesting that high social stress induces characteristic expression patterns in both the upper face and mouth region.

\begin{table}[tb]
\centering
\caption{Top-5 frequency rankings of facial action units (AUs) in the social press task.}
\label{tab:social_pressor_au_ranking}
\begin{tabular}{cccc}
\toprule
\textbf{Rank} & \textbf{All AU} & \textbf{MaE AU)} & \textbf{ME AU)} \\
\midrule
1 & AU24 & AU24 & AU24 \\
2 & AU20 & AU2  & AU20 \\
3 & AU2  & AU1  & AU17 \\
4 & AU17 & AU5  & AU4  \\
5 & AU1  & AU20 & AU2  \\
\bottomrule
\end{tabular}
\end{table}

To further assess whether AU counts differed across social-stress levels, we conducted repeated-measures ANOVA (\cref{tab:stress_speech_au_combined}). The total number of AUs differed significantly across the three levels ($F(2,128)=19.680$, $p<0.001$, $\eta_p^2=0.235$). Significant differences were observed for AU2 ($F(2,128)=3.460$, $p<0.05$, $\eta_p^2=0.051$), AU4 ($F(2,128)=3.425$, $p<0.05$, $\eta_p^2=0.051$), AU5 ($F(2,128)=4.582$, $p<0.05$, $\eta_p^2=0.067$), AU17 ($F(2,128)=7.904$, $p<0.01$, $\eta_p^2=0.110$), AU20 ($F(2,128)=4.055$, $p<0.05$, $\eta_p^2=0.060$), and AU24 ($F(2,128)=11.880$, $p<0.001$, $\eta_p^2=0.157$). AU1 did not show a significant difference across the three conditions ($F(2,128)=2.099$, $p=0.127$, $\eta_p^2=0.032$). In addition, AU7 ($F(2,128)=3.212$, $p<0.05$, $\eta_p^2=0.048$), AU9 ($F(2,128)=5.584$, $p<0.01$, $\eta_p^2=0.080$), AU23 ($F(2,128)=6.256$, $p<0.01$, $\eta_p^2=0.089$), and AU63 ($F(2,128)=4.152$, $p<0.05$, $\eta_p^2=0.061$) also differed significantly across conditions.

These results indicate that social stress has a stronger and more specific influence on facial expressions than cognitive stress. In particular, both MaE counts and multiple AUs varied significantly across the three social-stress conditions, suggesting that speaking under evaluative or deceptive contexts induces more distinctive facial-expression patterns.

\begin{table}[tb]
\centering
\caption{Facial AU statistics across stress-speech conditions.}
\label{tab:stress_speech_au_combined}
\setlength{\tabcolsep}{5pt}
\resizebox{\textwidth}{!}{
\begin{tabular}{lccc ccc cc}
\toprule
& \multicolumn{3}{c}{\textbf{Count}} & \multicolumn{3}{c}{\textbf{Mean $\pm$ Std}} & \multirow{2}{*}{\textbf{F}} & \multirow{2}{*}{\boldmath$\eta_p^2$} \\
\cmidrule(lr){2-4} \cmidrule(lr){5-7}
\textbf{AU} & \textbf{Lying} & \textbf{Speech} & \textbf{Reading} & \textbf{Lying} & \textbf{Speech} & \textbf{Reading} &  &  \\
\midrule
AU1  & 127 & 120 & 66 & 2.89$\pm$3.41 & 2.73$\pm$3.99 & 1.53$\pm$2.47 & 2.099     & 0.032 \\
AU2  & 139 & 123 & 58 & 3.16$\pm$3.67 & 2.80$\pm$3.96 & 1.35$\pm$2.29 & 3.460*    & 0.051 \\
AU4  & 117 & 48  & 44 & 2.66$\pm$4.84 & 1.09$\pm$2.00 & 1.02$\pm$2.31 & 3.425*    & 0.051 \\
AU5  & 84  & 47  & 31 & 1.91$\pm$2.50 & 1.07$\pm$1.44 & 0.72$\pm$1.53 & 4.582*    & 0.067 \\
AU6  & 1   & 2   & 0  & 0.02$\pm$0.15 & 0.05$\pm$0.30 & 0.00$\pm$0.00 & 0.589     & 0.009 \\
AU7  & 72  & 28  & 18 & 1.64$\pm$3.65 & 0.64$\pm$1.71 & 0.42$\pm$0.91 & 3.212*    & 0.048 \\
AU9  & 14  & 3   & 1  & 0.32$\pm$0.67 & 0.07$\pm$0.33 & 0.02$\pm$0.15 & 5.584**   & 0.080 \\
AU10 & 14  & 8   & 3  & 0.32$\pm$0.74 & 0.18$\pm$0.50 & 0.07$\pm$0.26 & 2.338     & 0.035 \\
AU11 & 2   & 0   & 0  & 0.05$\pm$0.30 & 0.00$\pm$0.00 & 0.00$\pm$0.00 & 0.988     & 0.015 \\
AU12 & 19  & 15  & 5  & 0.43$\pm$1.42 & 0.34$\pm$1.01 & 0.12$\pm$0.54 & 1.024     & 0.016 \\
AU13 & 4   & 1   & 0  & 0.09$\pm$0.60 & 0.02$\pm$0.15 & 0.00$\pm$0.00 & 0.753     & 0.012 \\
AU14 & 69  & 54  & 22 & 1.57$\pm$2.84 & 1.23$\pm$2.42 & 0.51$\pm$0.99 & 2.527     & 0.038 \\
AU15 & 8   & 2   & 0  & 0.18$\pm$0.72 & 0.05$\pm$0.21 & 0.00$\pm$0.00 & 2.047     & 0.031 \\
AU16 & 7   & 0   & 0  & 0.16$\pm$0.61 & 0.00$\pm$0.00 & 0.00$\pm$0.00 & 2.980     & 0.044 \\
AU17 & 135 & 118 & 32 & 3.07$\pm$3.11 & 2.68$\pm$3.53 & 0.74$\pm$1.80 & 7.904**   & 0.110 \\
AU20 & 155 & 117 & 58 & 3.52$\pm$4.14 & 2.66$\pm$3.32 & 1.35$\pm$3.21 & 4.055*    & 0.060 \\
AU22 & 1   & 0   & 0  & 0.02$\pm$0.15 & 0.00$\pm$0.00 & 0.00$\pm$0.00 & 0.988     & 0.015 \\
AU23 & 52  & 35  & 13 & 1.18$\pm$1.44 & 0.80$\pm$0.60 & 0.30$\pm$0.60 & 6.256**   & 0.089 \\
AU24 & 243 & 180 & 67 & 5.52$\pm$4.80 & 4.09$\pm$3.83 & 1.56$\pm$2.50 & 11.880*** & 0.157 \\
AU25 & 3   & 2   & 0  & 0.07$\pm$0.45 & 0.05$\pm$0.21 & 0.00$\pm$0.00 & 0.625     & 0.010 \\
AU26 & 12  & 5   & 2  & 0.27$\pm$0.82 & 0.11$\pm$0.39 & 0.05$\pm$0.31 & 1.930     & 0.029 \\
AU27 & 5   & 0   & 0  & 0.11$\pm$0.54 & 0.00$\pm$0.00 & 0.00$\pm$0.00 & 1.943     & 0.029 \\
AU28 & 26  & 24  & 8  & 0.59$\pm$1.35 & 0.55$\pm$1.19 & 0.19$\pm$0.66 & 1.726     & 0.026 \\
AU43 & 2   & 1   & 0  & 0.05$\pm$0.21 & 0.02$\pm$0.15 & 0.00$\pm$0.00 & 0.996     & 0.015 \\
AU44 & 13  & 10  & 4  & 0.30$\pm$0.59 & 0.23$\pm$0.52 & 0.09$\pm$0.48 & 1.614     & 0.025 \\
AU62 & 1   & 0   & 0  & 0.02$\pm$0.15 & 0.00$\pm$0.00 & 0.00$\pm$0.00 & 0.988     & 0.015 \\
AU63 & 23  & 26  & 2  & 0.52$\pm$0.93 & 0.59$\pm$1.35 & 0.05$\pm$0.21 & 4.152*    & 0.061 \\
AU68 & 2   & 2   & 1  & 0.05$\pm$0.30 & 0.05$\pm$0.30 & 0.02$\pm$0.15 & 0.104     & 0.002 \\
\midrule
Total & -- & -- & -- & 30.68$\pm$17.27 & 22.07$\pm$16.68 & 10.12$\pm$11.29 & 19.680*** & 0.235 \\
\bottomrule
\end{tabular}}
\vspace{2mm}
\begin{minipage}{0.98\linewidth}
\footnotesize
\textit{Note:} * $p<0.05$, ** $p<0.01$, *** $p<0.001$.
\end{minipage}
\end{table}

\subsection{Discussion}
Overall, the statistical analyses support the validity of our multi-level stress-induction design from a dataset perspective. Cognitive stress mainly contributes subtle AU-level variation, especially in the upper face, whereas social stress yields more pronounced differences at both the expression-count level and the AU level. This complementary pattern is desirable for CDER-SME, because it increases behavioral diversity without reducing the setting to a single stress source. In other words, the two stress paradigms enrich the dataset in different but mutually supportive ways: cognitive stress contributes fine-grained facial tension cues, while social stress induces more discriminative and ecologically realistic expression changes. Together, they improve the utility of the dataset for studying subtle facial behavior, stress-related expression dynamics, and cross-modal Event-RGB ME analysis.

\section{AU-rule-based Emotion Label Inference}

To ensure objective and reproducible emotion annotation, we infer emotion labels from annotated facial action units (AUs) rather than from self-reported affect. This design follows the FACS-based objective coding protocol adopted in the main paper, where labels are grounded in observable facial muscle dynamics instead of subjective interpretation. In practice, each sample is first represented as a set of annotated AUs, and laterality prefixes (e.g., left/right markers) are removed so that the emotion inference depends on the underlying facial action configuration rather than side-specific notation.

As detailed in \cref{alg:au_mapping}, we implement a priority-based hierarchical matching procedure. Because spontaneous MEs are often subtle, incomplete, or blended, a flat one-shot mapping may produce ambiguous assignments when multiple AU patterns partially overlap. We therefore prioritize emotions with relatively distinctive AU signatures~\cite{dong2022spontaneous}. Concretely, the mapping follows the descending priority order used in the main paper: Happiness if {12, 6, 28} \(\cap A \neq \emptyset\); Surprise if \(2 \in A\); Anger if {16, 22, 23} \(\cap A \neq \emptyset\); Fear if {1, 4, 5, 25} \(\subseteq A\); Disgust if {7, 24} \(\cap A \neq \emptyset\); and Sadness if ({4, 5} \(\subseteq A\)) or \(43 \in A\). Samples that do not satisfy any prototypical rule are assigned to Others. This hierarchical design ensures that labels remain deterministic and directly traceable to AU evidence.

The rationale for this design is twofold. First, it avoids dependence on subjective reports, which can be unreliable in high-pressure or deceptive settings. Second, it provides a transparent labeling protocol that can be reproduced or audited by future users of the dataset. For example, a sample containing AU2 is assigned to Surprise because AU2 is treated as a distinctive marker under the current hierarchy, while a sample containing AU7 and/or AU24 is assigned to Disgust according to the corresponding rule. By contrast, samples with partial, weak, or blended AU configurations that do not meet any prototypical criterion are conservatively grouped into Others. In this way, the final emotion labels function as AU-derived objective categories for benchmarking, rather than as claims about the subjects’ full internal affective states.

\begin{figure}[tb]
  \centering
  \begin{subfigure}{0.48\linewidth}
    \includegraphics[width=\linewidth]{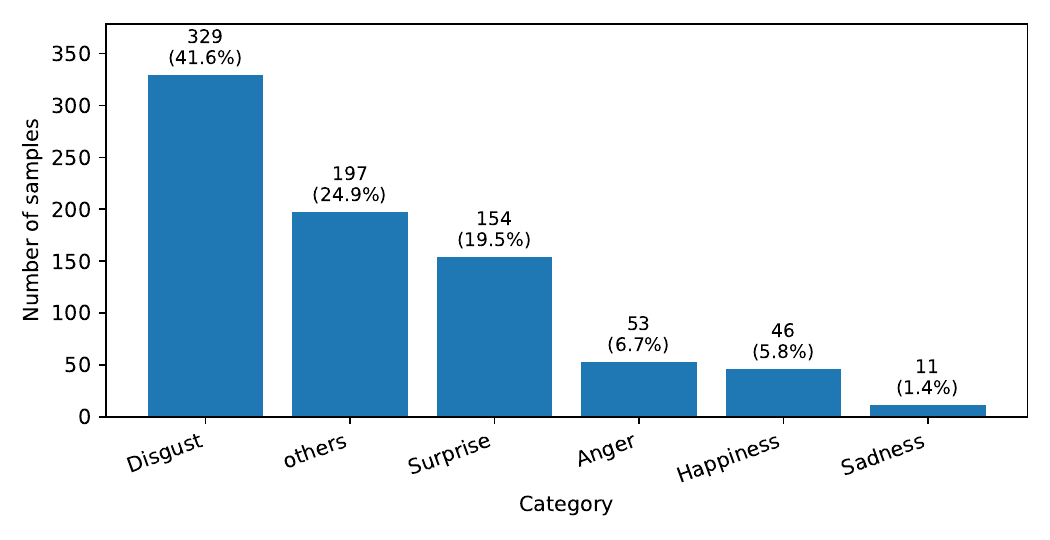}
    \caption{790 raw pairs}
    \label{fig:790raw}
  \end{subfigure}
  \hfill
  \begin{subfigure}{0.48\linewidth}
    \includegraphics[width=\linewidth]{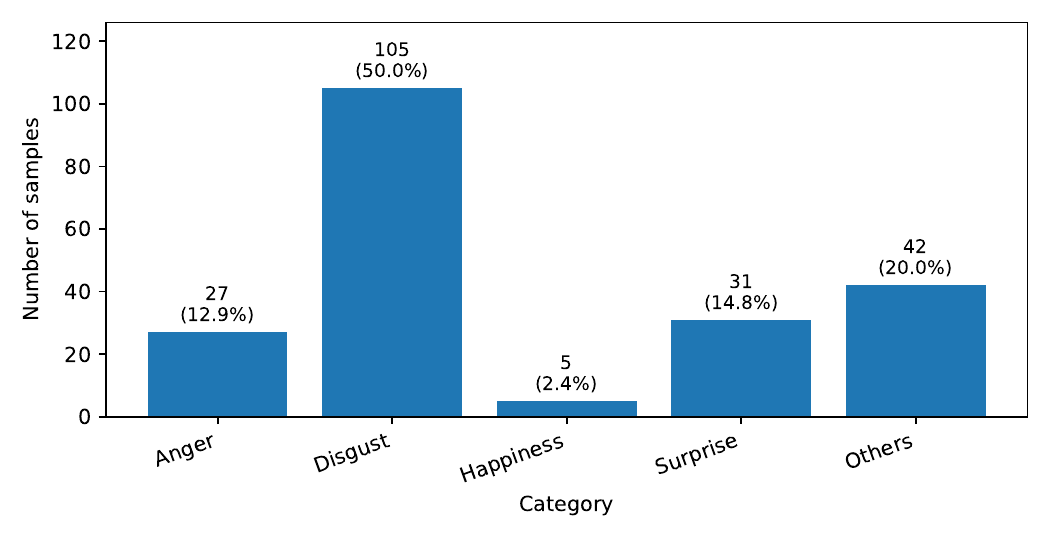}
    \caption{210 aligned pairs}
    \label{fig:210aligned}
  \end{subfigure}
  \caption{Emotion class distributions of the 790 raw Event–RGB pairs and the 210 aligned Event–RGB pairs. Emotion labels are derived from AU-rule-based objective inference.}
  \label{fig:emo_ditribution}
\end{figure}

\cref{fig:emo_ditribution} shows the resulting emotion distributions for the 790 raw Event–RGB pairs and the 210 aligned pairs. In the raw 790-pair pool, the categories are Disgust, Others, Surprise, Anger, Happiness, and Sadness, whereas the final 210-pair aligned benchmark contains Anger, Disgust, Happiness, Surprise, and Others. This difference reflects the interaction between AU-rule-based labeling and the subsequent quality-control filtering of aligned samples, rather than a change in the annotation protocol itself. In particular, no sample in the current benchmark satisfies the Fear rule after final curation, so Fear is merged into Others, consistent with the main paper. Likewise, some low-frequency classes in the raw pool do not remain in the aligned benchmark after strict synchronization and registration filtering.

Overall, this AU-rule-based inference protocol provides an objective and reproducible bridge from expert AU annotations to benchmark-level emotion categories. It is therefore well suited to CDER-SME, whose primary goal is to support fair comparison and cross-modal ME analysis under a realistic stress-induction setting.

\begin{algorithm}[tb]
\caption{Priority-based Hierarchical Emotion Mapping.}\label{alg:au_mapping}
\KwIn{Raw AU set $\mathcal{A}_{raw}$ with laterality prefixes.}
\KwOut{Emotion label $y \in \{\text{Happiness, Surprise, Anger, Fear, Disgust, Sadness, Others}\}$.}
\BlankLine\tcp{Standardize AU indices by removing 'L'/'R' prefixes}$\mathcal{A} \leftarrow \{ \text{index}(a) \mid a \in \mathcal{A}_{raw} \}$;
\BlankLine\tcp{Hierarchical Inference based on FACS logic}\uIf{$\{12, 6, 28\} \cap \mathcal{A} \neq \emptyset$}{\Return \textbf{Happiness};}
\uElseIf{$2 \in \mathcal{A}$}{\Return \textbf{Surprise};}
\uElseIf{$\{16, 22, 23\} \cap \mathcal{A} \neq \emptyset$}{\Return \textbf{Anger};}
\uElseIf{$\{1, 4, 5, 25\} \subseteq \mathcal{A}$}{\Return \textbf{Fear};}
\uElseIf{$\{7, 24\} \cap \mathcal{A} \neq \emptyset$}{\Return \textbf{Disgust};}
\uElseIf{$(\{4, 5\} \subseteq \mathcal{A}) \lor (43 \in \mathcal{A})$}{\Return \textbf{Sadness};}
\Else{\Return \textbf{Others};}
\end{algorithm}

\section{Quality-Control Filtering of the 790 Raw Event–RGB Pairs}

To establish a reliable benchmark core for cross-modal  analysis, we applied a strict quality-control filtering process to the 790 raw Event–RGB pairs. The purpose of this step is not merely to discard imperfect recordings, but to separate a high-fidelity aligned subset for standardized benchmarking from a broader raw-pair pool that still reflects realistic cross-device acquisition noise. The resulting sample accounting is summarized in \cref{tab:data_stats}.

Among the 790 raw pairs, 210 samples were retained as valid aligned pairs, while the remaining 580 samples were excluded due to identifiable quality-control failures. Specifically, the excluded cases consist of Nixie Tube Occlusion / Offset (121), Temporal Misalignment caused by initial overexposure (270), Incomplete Facial Capture (5), Recording Failure with missing event data (76), Synchronization Threshold Outliers (69), and Incomplete Event Sequences caused by acquisition failure (39). These categories sum exactly to the full 790-pair raw bimodal pool and provide a transparent bookkeeping of the filtering process.

This filtering protocol is consistent with the benchmark design in the main paper. In particular, the dominant failure modes are hardware- and acquisition-related rather than behavior-related, including overexposure-induced temporal errors, digital marker instability, and event-stream acquisition failures. As a result, the retained 210 samples form a high-confidence aligned benchmark subset, while the full 790-pair raw pool remains valuable as a realistic resource for future research on weak synchronization, robust alignment, event generation, and domain adaptation under sensor noise.

Importantly, the filtering process should not be interpreted as redefining the dataset itself, but rather as organizing CDER-SME into different utility tiers. The 210 retained pairs correspond to the high-fidelity benchmark core for precise cross-modal evaluation, whereas the 790 raw pairs preserve more challenging real-world imperfections. This distinction is especially important in the decoupled cross-device setting, where practical deployment conditions inevitably introduce synchronization and registration errors that are absent in rigidly coupled laboratory systems.

To assess whether the filtering process introduced undesirable behavioral distortion, we further analyze the representativeness of the retained subset in \cref{sec:selection_bias}. As shown there, the selected 210 samples remain statistically consistent with the original 790-pair pool with respect to the key expression-related variables, supporting their use as the official aligned benchmark.

\begin{table}[tb]
\centering
\caption{Detailed sample accounting of the quality-control filtering process applied to the 790 raw Event–RGB pairs.}
\label{tab:data_stats}
\begin{tabular}{lc}
\toprule
\textbf{Category} & \textbf{Count} \\
\midrule
Valid Samples & 210 \\
Nixie Tube Occlusion / Offset & 121 \\
Temporal Misalignment (Initial Overexposure) & 270 \\
Incomplete Facial Capture & 5 \\
Recording Failure (No Event Data) & 76 \\
Synchronization Threshold Outlier & 69 \\
Incomplete Event Sequences (Acquisition Failure) & 39 \\
\midrule
\textbf{Total} & \textbf{790} \\
\bottomrule
\end{tabular}
\end{table}
\par

No significant differences were observed in the distribution of expression categories ($p=0.163$) or duration bins ($p=0.051$). These results indicate that the filtered subset maintains the core behavioral characteristics of the original data without systematic shifts. While a significant difference exists in the experimental task distribution ($p < 0.001$), this discrepancy is primarily attributed to hardware-related factors during the quality control process, such as Display tube occlusions or sensor acquisition failures in specific recording sessions. Crucially, these hardware issues are independent of the subjects' facial behaviors. Thus, the 210-sample subset serves as a statistically sound representative for benchmarking cross-modal ME analysis.

\section{Analysis of Potential Selection Bias After Sample Filtering}
\label{sec:selection_bias}

To assess whether the quality-control filtering from 790 raw Event–RGB pairs to 210 aligned pairs introduced undesirable selection bias, we compared the retained subset with the original raw pool along three dimensions: experiment type, expression type (ME vs. MaE), and expression duration. For each dimension, a chi-square (\(\chi^2\)) test was performed to evaluate whether the retained subset deviated significantly from the original distribution (\cref{tab:selection_bias_combined}).

The results show that the aligned subset remains statistically consistent with the original 790-pair pool in terms of the two core expression-related variables. No significant difference is observed for expression type (\(\chi^2=1.945, p=0.163080\)), and the duration-bin distribution is likewise not significantly altered (\(\chi^2=7.750, p=0.051476\)). These findings indicate that the retained 210 samples preserve the essential behavioral structure of the original bimodal dataset with respect to ME/MaE composition and temporal extent.

A significant difference does appear in the experiment-type distribution (\(\chi^2=33.146, p=0.000004\)). In particular, the relative proportion of Social Stress (Lying) increases after filtering, while several cognitive-stress categories decrease. However, this shift should not be interpreted as evidence of behavior-driven sampling bias. Instead, it is primarily explained by hardware- and acquisition-related failure modes during the quality-control process, including temporal misalignment caused by initial overexposure, marker occlusion/offset, and event-stream recording failures. Because these issues arise from sensor and recording conditions rather than facial-expression content, the resulting task-type skew is better understood as a byproduct of cross-device acquisition constraints. 

Accordingly, we do not claim that the aligned subset is perfectly representative of the original raw-pair pool in every respect. Rather, our conclusion is that the retained subset remains representative with respect to the key variables most relevant for expression benchmarking, while task-condition proportions are only partially preserved after filtering. This distinction is important for interpreting the benchmark: Tier-3 is well suited for high-fidelity cross-modal evaluation, whereas Tier-2 retains broader task diversity under realistic synchronization and acquisition noise.

Overall, these analyses support the use of the 210 aligned pairs as the official benchmark core of CDER-SME. At the same time, the full 790-pair raw pool remains an important complementary resource for future work on alignment robustness, weak synchronization, and cross-device multimodal learning.

\begin{table}[tb]
\centering
\caption{Selection-bias analysis before and after quality-control filtering. The table compares the distributions of experiment type, expression type, and expression duration between the original 790 raw Event–RGB pairs and the retained 210 aligned pairs, together with the corresponding chi-square test results.}
\label{tab:selection_bias_combined}
\setlength{\tabcolsep}{6pt}
\resizebox{\textwidth}{!}{
\begin{tabular}{llcccc}
\toprule
\textbf{Dimension} & \textbf{Category} & \textbf{790 samples (\%)} & \textbf{210 samples (\%)} & \boldmath$\chi^2$ & \textbf{\emph{p}-value} \\
\midrule
\multirow{6}{*}{Experiment type}
& Cognitive Stress (HS) & 158 (20.0\%) & 21 (10.0\%)  & \multirow{6}{*}{33.146} & \multirow{6}{*}{0.000004} \\
& Cognitive Stress (LS) & 92 (11.6\%)  & 13 (6.2\%)   &  &  \\
& Cognitive Stress (NS) & 103 (13.0\%) & 17 (8.1\%)   &  &  \\
& Social Stress (Reading) & 49 (6.2\%)  & 15 (7.1\%)   &  &  \\
& Social Stress (Speech)  & 151 (19.1\%) & 44 (21.0\%) &  &  \\
& Social Stress (Lying)   & 237 (30.0\%) & 100 (47.6\%) &  &  \\
\midrule
\multirow{2}{*}{Expression type}
& ME  & 395 (50.0\%) & 117 (55.7\%) & \multirow{2}{*}{1.945} & \multirow{2}{*}{0.163080} \\
& MaE & 395 (50.0\%) & 93 (44.3\%)  &  &  \\
\midrule
\multirow{4}{*}{Duration bin}
& Very short ($0$--$0.3$s) & 143 (18.1\%) & 37 (17.6\%)  & \multirow{4}{*}{7.750} & \multirow{4}{*}{0.051476} \\
& Short ($0.3$--$0.6$s)    & 337 (42.7\%) & 110 (52.4\%) &  &  \\
& Medium ($0.6$--$1$s)     & 167 (21.1\%) & 37 (17.6\%)  &  &  \\
& Long ($>1$s)             & 143 (18.1\%) & 26 (12.4\%)  &  &  \\
\bottomrule
\end{tabular}}
\end{table}

These findings indicate that the filtered subset preserves the essential behavioral structure of the original dataset and is therefore suitable for downstream analysis.

\section{Detailed Temporal Alignment: Formalization and Metrics}

Given the absence of external ground-truth (GT) for cross-modal synchronization in a decoupled real-world acquisition setup, we evaluate alignment quality through temporal consistency between the RGB stream and the event stream. The core assumption is that, if the two modalities are well aligned, the intensity of motion/change within the same expression segment should exhibit similar temporal trends. This criterion is consistent with the benchmark-quality assessment in the main paper, where alignment reliability is established through quantitative cross-modal consistency rather than rigid hardware-level coaxial synchronization.

\subsection{Motion Profile Construction}

For each annotated segment $[t_s, t_e]$, we construct one-dimensional motion profiles for both modalities.

\paragraph{RGB Motion Curve ($m_R(t)$).}
For an RGB frame sequence $\{I_t\}$, the motion intensity is defined as the mean absolute difference between consecutive frames:
\begin{equation}
m_R(t) = \frac{1}{HW} \sum_{x,y} \left| I_t(x,y) - I_{t-1}(x,y) \right|,
\end{equation}
where $H$ and $W$ denote the frame resolution. This curve captures the temporal development of facial motion, including onset-to-apex fluctuations relevant to ME dynamics.

\paragraph{Event Intensity Curve ($m_E(t)$).}
Within the same temporal segment, the event stream is partitioned into $B = 120$ uniform time bins. The event intensity in bin $k$ is defined as
\begin{equation}
m_{E,k} = \mathrm{count}\{e_i \mid t_i \in \mathrm{bin}_k\}, \quad k = 1, \ldots, B.
\end{equation}
This profile reflects the activity intensity captured by the event sensor and summarizes the temporal distribution of asynchronous brightness changes.

\subsection{Correlation Scanning and Optimal Lag}

To quantify alignment, both curves are resampled to a common length $L$ ($30 \leq L \leq 120$) and normalized via z-score to eliminate differences in magnitude. Let $r_t$ and $e_t$ denote the standardized RGB and event curves, respectively. We then perform a temporal-shift correlation scan within a discrete shift range $l \in [-L_{\max}, L_{\max}]$, where $L_{\max}$ corresponds to $\pm 20$ bins:
\begin{equation}
c_l = \mathrm{corr}(r_t, e_{t+l}).
\end{equation}

The optimal temporal lag $\delta^*$ and maximum correlation coefficient $c^*$ are determined by
\begin{equation}
\delta^* = \arg\max_l c_l, \qquad c^* = c_{\delta^*}.
\end{equation}

For interpretability, $\delta^*$ is further converted into a physical time offset $\Delta t^*$ (in seconds):
\begin{equation}
\Delta t^* = \frac{\delta^*}{L - 1} \times \mathrm{duration},
\end{equation}
where $\mathrm{duration}$ denotes the temporal length of the RGB segment. Intuitively, a high $c^*$ achieved together with a small $|\Delta t^*|$ indicates that the two modalities exhibit highly consistent temporal evolution without requiring substantial temporal shifting.

\subsection{Peak Distance Ratio (PDR)}

To complement the correlation-based measure, we introduce the Peak Distance Ratio (PDR) to evaluate whether the strongest activity moment occurs at a similar temporal location in both modalities. Let $P_R$ and $P_E$ denote the indices of the peaks of the RGB and event curves:
\begin{equation}
P_R = \arg\max_t r_t, \qquad P_E = \arg\max_t e_t.
\end{equation}

The PDR is defined as
\begin{equation}
\mathrm{PDR} = \frac{|P_R - P_E|}{L - 1}.
\end{equation}

A low PDR indicates that the dominant motion peaks of the two modalities are temporally consistent. In practice, $c^*$, $|\Delta t^*|$, and PDR capture complementary aspects of alignment quality: $c^*$ measures overall trend consistency, $|\Delta t^*|$ measures the amount of temporal adjustment needed for best correspondence, and PDR measures whether the most salient activity moment is synchronized across modalities.

\subsection{Detailed Statistical Results}

We evaluated all 210 aligned Event--RGB pairs using the metrics above. The median maximum correlation $c^*$ is $0.6222$, and the 95th percentile reaches $0.8848$, indicating high temporal consistency for the vast majority of the aligned benchmark samples. The median optimal physical lag $|\Delta t^*|$ is restricted to $0.0831\,\mathrm{s}$, showing that only a small residual offset remains after alignment. In addition, the median PDR is $0.1471$, confirming that the strongest motion moments are generally well synchronized across modalities.

Overall, these statistics support the reliability of the retained Tier-3 aligned subset as the official benchmark core of CDER-SME. Importantly, because the dataset is collected under a non-coaxial, non-pre-calibrated, decoupled cross-device setting, such quantitative consistency is particularly meaningful: it demonstrates that high-fidelity cross-modal correspondence can still be achieved under realistic deployment constraints through algorithmic alignment rather than rigid hardware coupling.

\section{Reference Baseline Architecture Details}

This section documents the reference baseline released with CDER-SME, whose role is to facilitate reproducible benchmarking rather than to serve as the main contribution of the paper.

\subsubsection{Event-Temporal Multi-scale Aggregation (E-TMA)}
\label{sec:etma}

MEs are characterized by brief, weak, and highly localized facial motions whose dynamics unfold across multiple temporal scales, from a gradual onset to an abrupt impulse around the apex. For event cameras, this creates a fundamental tension: at microsecond resolution, the asynchronous event stream is highly informative yet often sparse and semantically under-specified when naively discretized. To bridge this sparsity--semantics gap and explicitly model motion boundaries, we introduce \textbf{Event-Temporal Multi-scale Aggregation (E-TMA)}, which aggregates event evidence over physically meaningful, asymmetric temporal windows anchored at the apex.

\paragraph{Structured Temporal Partitioning}
Instead of blindly stacking fixed-length event frames, we exploit the kinematics of facial muscle activation and use the apex time $t_{apex}$ as an anchor. Given an event stream $\mathcal{E}$, we partition it into three temporal spans with increasing temporal horizons:
\begin{equation}
\mathcal{T}_{long} = [t_{onset},\, t_{apex}], \quad
\mathcal{T}_{mid} = [t_{apex}-2\Delta t,\, t_{apex}], \quad
\mathcal{T}_{short} = [t_{apex}-\Delta t,\, t_{apex}].
\end{equation}
where $\Delta t$ denotes the duration of one RGB frame. The long-range window captures the complete trajectory from quiescence to peak contraction, the mid-range window emphasizes acceleration changes immediately before the peak, and the short-range window captures the strongest impulsive signals at the moment of peak activation. This asymmetric design reflects that discriminative ME cues often concentrate near the apex, while the longer horizon provides contextual evolution that helps disambiguate noise and spurious motion.

\paragraph{Scale-Aware Feature Aggregation}
For each temporal span $\mathcal{T}_i$, we voxelize the corresponding sub-stream $\mathcal{E}_{\mathcal{T}_i}$ and map it into a latent semantic space using a shared projection $\Phi_{shared}(\cdot)$. We then apply scale-dependent pooling whose receptive field increases with window length: longer windows benefit from larger spatial integration to suppress accumulated sensor noise, whereas shorter windows require small kernels to preserve sharp motion boundaries. Formally,
\begin{equation}
e'_{i} = \mathcal{P}_{s_i}\Big(\Phi_{shared}(\mathrm{Voxel}(\mathcal{E}_{\mathcal{T}_i}))\Big), 
\quad i \in \{short,\, mid,\, long\},
\label{eq:etma_scale_pool}
\end{equation}
where $\mathcal{P}_{s_i}$ denotes a pooling operator whose stride and kernel size are monotonically increased for longer durations (e.g., a larger $(n{+}1)\!\times\!(n{+}1)$ kernel for $\mathcal{T}_{long}$ to reduce redundancy and noise, and a $2\!\times\!2$ kernel for $\mathcal{T}_{short}$ to retain fine-grained edges and impulses).

\paragraph{Motion Boundary Reinforcement}
Finally, the multi-scale features $\{e'_{short}, e'_{mid}, e'_{long}\}$ are fused via \textit{Concat--Conv} to perform cross-scale aggregation from coarse semantic evolution to fine motion transients. We concatenate features along the channel dimension and refine them with a lightweight bottleneck (e.g., a 3D bottleneck block) to obtain a unified event representation:
\begin{equation}
f_e = \mathrm{Bottleneck}\Big(\mathrm{Concat}(e'_{short},\, e'_{mid},\, e'_{long})\Big).
\label{eq:etma_fuse}
\end{equation}
This design enables the network to capture both (i) the global evolution trend from onset to apex and (ii) the subtle, $\Delta t$-level motion discontinuities near the apex, resulting in a high-discriminability event feature robust to sparsity and background redundancy.

\subsubsection{Shallow Triple Stream Spatio-Temporal Network (STSTNet)}

For the RGB modality, we utilize a three-stream architecture optimized for the low-intensity motion fields of MEs. We extract the horizontal flow $u$, vertical flow $v$, and optical strain $\varepsilon$ between the onset and apex frames. These fields serve as inputs to a parallel 3D-CNN bank:
$$f_{rgb} = \text{Concat}(\text{Conv}_{3\times3\times3}(\phi)) \text{ where } \phi \in \{u, v, \varepsilon\}$$By specifically targeting optical strain, our model emphasizes local skin deformations that are often overlooked by standard CNNs, providing a robust spatial anchor for the event modality.

\subsubsection{Second-Order Cross-Modal Consistency (CMC)}
\label{sec:cmc}

While RGB frames provide rich appearance cues, event streams emphasize motion-induced intensity changes and are less sensitive to illumination and blur. This inherent modality gap often leads to feature misalignment, especially under subject-specific appearance variations. To explicitly couple the two branches at a distributional level, we introduce \textbf{Second-Order Cross-Modal Consistency (CMC)}, which aligns RGB and event representations via their \emph{second-order statistics}. Compared to first-order matching, covariance captures channel-wise co-activation patterns and encourages the two modalities to share consistent correlation structure, which empirically improves class separability and robustness.

\paragraph{Batch-wise empirical covariance}
Let $\{f_i^{\mathrm{RGB}}\}_{i=1}^{N}$ and $\{f_i^{\mathrm{Event}}\}_{i=1}^{N}$ denote the per-sample features from the RGB branch (e.g., STSNet) and the event branch (e.g., E-TMA), respectively, within a mini-batch of size $N$. We compute the batch means
\begin{equation}
\mu^{\mathrm{RGB}}=\frac{1}{N}\sum_{i=1}^{N} f_i^{\mathrm{RGB}}, \qquad
\mu^{\mathrm{Event}}=\frac{1}{N}\sum_{i=1}^{N} f_i^{\mathrm{Event}},
\end{equation}
and then form the empirical covariance matrices
\begin{equation}
\begin{split}
\Sigma^{\mathrm{RGB}} &= \frac{1}{N-1}\sum_{i=1}^{N}\left(f_i^{\mathrm{RGB}}-\mu^{\mathrm{RGB}}\right)\left(f_i^{\mathrm{RGB}}-\mu^{\mathrm{RGB}}\right)^{\top}, \\
\Sigma^{\mathrm{Event}} &= \frac{1}{N-1}\sum_{i=1}^{N}\left(f_i^{\mathrm{Event}}-\mu^{\mathrm{Event}}\right)\left(f_i^{\mathrm{Event}}-\mu^{\mathrm{Event}}\right)^{\top}.
\end{split}
\label{eq:cmc_cov}
\end{equation}
Here, $\Sigma$ characterizes channel-to-channel dependencies, reflecting how discriminative micro-motion/appearance cues co-vary across feature dimensions.

\paragraph{Second-order alignment objective}
We enforce cross-modal consistency by minimizing the Frobenius distance between the two covariance matrices:
\begin{equation}
\mathcal{L}_{\mathrm{cov}}=\left\|\Sigma^{\mathrm{RGB}}-\Sigma^{\mathrm{Event}}\right\|_{F}^{2}.
\label{eq:cmc_loss}
\end{equation}
Optimizing Eq.~\eqref{eq:cmc_loss} aligns RGB and event features at the level of second-order structure, encouraging both modalities to encode similar correlation patterns for emotion-relevant dynamics. Intuitively, this regularizer promotes a form of \emph{statistical ``whitening''} against nuisance factors: subject identity and illumination variations tend to manifest as modality- or instance-specific mean/texture biases, whereas ME cues are better captured by consistent co-activation relations across channels. Consequently, CMC improves robustness to lighting changes and sample-specific artifacts while strengthening cross-modal discriminability.

\paragraph{Integration with the overall pipeline}
CMC is applied to the intermediate features produced by the RGB and event encoders before cross-modal interaction modules (i.e., CrossMamba) and the final classifier.

\subsubsection{Bi-Directional CrossMamba Fusion}

To achieve deep interaction between the spatial-heavy RGB and temporal-heavy Event features, we move beyond vanilla attention and propose a Bi-Directional Cross-Selective State Space Model (Bi-CS-SSM). The core innovation lies in the cross-modal hidden attention matrix $\alpha_{i,j}$ derived from the Mamba selection mechanism. We define the CrossMamba operation where the query $Q$ is derived from one modality $x_q$ and the keys/values $(K, V)$ from the other $x_v$:
$$y_i = \sum_{j=1}^{i} Q_i K_j H_{i,j} V_j$$
where $H_{i,j}$ represents the accumulated state transition from index $j$ to $i$. In paticular, 
\begin{itemize}
    \item for RGB-guided event selection, $x_q = f_{rgb}$ filters out sensor noise in the event stream by highlighting regions of significant optical flow;
    \item  for event-guided RGB enhancement, $x_v = f_{e}$ provides high-frequency motion cues to localize transient ROI changes within the RGB frames.
\end{itemize}
 The final feature is a summation of the causal and flipped non-causal scans, ensuring a global receptive field over the entire ME sequence.

\subsubsection{Joint Optimization}
The network is trained end-to-end using a weighted loss function that balances task-specific classification and cross-modal structural alignment:
$$\mathcal{L} = \alpha \mathcal{L}_{cls} + \beta \mathcal{L}_{cov}$$
where $\mathcal{L}_{cls} = CE(y, \hat{y})$, $\alpha$ and $\beta$ are hyperparameters, with optimal configuration defined based on the experiment result in \cref{subsec:hyperPara}.

\section{Official Evaluation Protocol}
We define the Tier-3 aligned subset as the primary benchmark of CDER-SME, because it provides the highest-confidence cross-modal synchronization and spatial consistency among all released Event–RGB pairs. The aligned subset contains 210 samples from 14 subjects, and is therefore the most suitable evaluation target for cross-modal MER under the decoupled acquisition setting. The main paper already positions Tier-3 as the benchmark core, while Tier-2 is released as a more challenging but noisier resource for robust alignment and adaptation research.

Following standard MER evaluation practice, we adopt a Leave-One-Subject-Out (LOSO) protocol to ensure subject-independent testing. In each fold, all samples from one subject are held out for testing, while the remaining subjects are used for training. This protocol prevents identity leakage and provides a fair estimate of cross-subject generalization. Performance is reported using Accuracy (ACC), Unweighted F1-score (UF1), and Unweighted Average Recall (UAR), as stated in the main paper. We emphasize UF1 and UAR because the class distribution is naturally imbalanced under the stress-induction paradigm and the aligned benchmark subset does not enforce artificial class balancing.

To ensure fair and reproducible comparison, we will release the official split files and evaluation scripts with the benchmark package. For multimodal methods evaluated on Tier-3, predictions must be made at the sample level under the provided subject-independent folds. Methods may use the released temporal annotations and benchmark metadata that are part of the official package, but no identity-overlapping samples are allowed across training and test folds. In addition, we recommend reporting both unimodal baselines (RGB-only and Event-only) and multimodal baselines, since the benchmark is specifically designed to evaluate modality complementarity under realistic cross-device conditions. The main paper shows that the two modalities provide complementary cues, with multimodal fusion outperforming single-modality baselines on the aligned benchmark.

Finally, we stress that Tier-2 and Tier-3 serve different evaluation purposes. Tier-3 should be regarded as the official benchmark for high-precision cross-modal MER, whereas Tier-2 is better interpreted as an auxiliary testbed for alignment robustness, weak synchronization, synthetic-event research, or domain adaptation. This distinction helps avoid conflating alignment quality with recognition performance and reflects the practical design philosophy of CDER-SME.

\section{Implementation Details}

All experiments were conducted on a workstation equipped with an NVIDIA GeForce RTX 4090 GPU with NVIDIA driver version 570.211.01. The software environment was based on Python 3.10.14, PyTorch 2.2.2+cu121, and CUDA 12.1.

Unless otherwise specified, the training configuration was set as follows: the number of training epochs was 800 for the released multimodal baseline, the batch size was 4, the initial learning rate was $1\times10^{-4}$, and the model dimension was $d_{\text{model}}=256$.

For modality-specific inputs, the event camera data with spatial resolution $640\times480$ were fed into the E-TMA module. For the RGB modality, optical flow was first computed from RGB frames, and the resulting flow representation was normalized and resized to $28\times28$ before being fed into STSNet.

During the CrossMamba training stage, the dual-modal inputs were configured as follows. The RGB feature input had shape $(1,16,5,5)$, corresponding to a flattened dimension of $16\times5\times5=400$. The event feature input had shape $(1,64,72,72)$, corresponding to a flattened dimension of $64\times72\times72=331{,}776$. Here, the leading dimension $1$ denotes the single-sample dimension after preprocessing.

Official configuration files and evaluation scripts will be released with the benchmark package.

\section{Reference Baseline on the Raw Bimodal Pool with Generated Events}

In addition to the official Tier-3 aligned benchmark, we provide an auxiliary reference evaluation on the larger raw bimodal pool to illustrate the feasibility of cross-modal learning under weaker synchronization conditions. These results are supplementary reference evaluations and should not be interpreted as replacing the official Tier-3 benchmark.

Since the 790 raw Event--RGB pairs are not all suitable for direct high-fidelity cross-modal benchmarking, we generate event representations from the RGB modality and use them only for supplementary reference experiments rather than as part of the primary benchmark. Specifically, we employed the DVS-Voltmeter event simulator~\cite{lin2022dvs} to synthesize event streams from high-frame-rate RGB videos. Given an RGB video clip, we first converted it into a temporally ordered sequence of image frames and assigned each frame a precise timestamp at microsecond resolution. The frames were then processed sequentially by the simulator. Based on the brightness changes between consecutive frames, DVS-Voltmeter estimates the event triggering probability and the temporal distribution of events using a statistical model based on stochastic processes, namely Brownian motion with drift, and finally generates a synthetic event stream.

Among the 790 raw RGB samples, 13 samples failed during event-generation preprocessing, leaving 777 valid samples for this auxiliary evaluation. We emphasize that this experiment should not be interpreted as the official benchmark result of CDER-SME. Instead, it serves as a reference baseline on a larger but less strictly curated sample pool, complementing the high-fidelity evaluation on the aligned subset.

Tab.~\ref{tab:generated_event_baseline} reports the results of our released reference baseline on the 777 valid samples. The model achieves an ACC of 0.6963, a UF1 of 0.5125, and a UAR of 0.5007. These results suggest that useful cross-modal supervisory signals can still be exploited on the larger raw sample pool when generated-event representations are used. At the same time, the evaluation protocol and data fidelity in this setting differ from those of the official aligned benchmark, and the corresponding results should therefore be interpreted as an auxiliary reference rather than as the main performance indicator of the dataset.

Overall, this experiment highlights the complementary role of the raw bimodal pool in CDER-SME. While Tier-3 remains the official benchmark core for precise Event--RGB evaluation, the larger raw pool can still support broader exploratory studies such as generated-event learning, weak synchronization, and robustness-oriented multimodal training.

\begin{table}[tb]
\centering
\caption{Reference baseline results on the Raw Bimodal Pool. }
\label{tab:generated_event_baseline}
\begin{tabular}{lccc}
\toprule
Method & ACC & UF1 & UAR \\
\midrule
Ours baseline on 790 raw samples & 0.6963 & 0.5125 & 0.5007 \\
\bottomrule
\end{tabular}

\end{table}

\section{Comparison with Single-Modality Reference Baselines}

To further demonstrate the utility of CDER-SME as a multimodal benchmark, we compare our released reference baseline against several representative single-modality baselines on the 210-sample aligned subset. This experiment is designed to assess how much complementary information can be obtained from RGB and event modalities individually, and whether the proposed benchmark setting is sufficiently challenging for both conventional frame-based and event-based pipelines. These results are supplementary reference evaluations and should not be interpreted as replacing the official Tier-3 benchmark.

For the RGB modality, we compare against three representative methods with open source code, i.e., MMNET~\cite{li2022mmnet}, BDCNN~\cite{chen2022block}, and HTNet~\cite{wang2024htnet}. For the event modality, we compare against two standard event representations, namely Voxel + CNN and Event Frame + CNN. To reduce the influence of training-length discrepancies and to avoid excessive overfitting in this relatively small benchmark, the CNN-based baselines are trained with practical schedules that ensure stable optimization. In particular, although some original training plans considered longer schedules, the corresponding loss curves had already reached stable regimes under the adopted settings, and the reported results are therefore considered sufficiently representative for benchmark comparison.

The results are summarized in \cref{tab:single_modal_comparison}. On the RGB branch, our reference baseline achieves the best performance among all compared methods, with an ACC of 0.5762, a UF1 of 0.3507, and a UAR of 0.3443. In comparison, MMNET obtains 0.4920/0.2509/0.2409, BDCNN obtains 0.4976/0.1329/0.1329, and HTNet obtains 0.3667/0.1872/0.1944 in terms of ACC/UF1/UAR. These results suggest that the aligned subset remains challenging for RGB-only recognition and that stronger temporal modeling and cross-sample feature aggregation are still beneficial even in the single-modality setting.

A similar trend is observed on the event branch. Our method achieves the strongest event-only performance, with an ACC of 0.5619, a UF1 of 0.3180, and a UAR of 0.3167, outperforming both Voxel + CNN (0.3667/0.3232/0.3093) and Event Frame + CNN (0.3524/0.2486/0.2301). Notably, while Voxel + CNN attains a UF1 close to that of our method, its lower ACC and UAR indicate less balanced overall recognition behavior across classes. This again supports the view that the event modality contains informative but challenging facial dynamics, and that CDER-SME provides a meaningful benchmark for evaluating event-based MER.

Overall, these single-modality comparisons support two conclusions. First, both RGB and event modalities are individually non-trivial on the aligned benchmark, confirming the difficulty of the dataset. Second, the consistent improvements of our reference baseline over the unimodal competitors indicate that CDER-SME is well suited for studying complementary multimodal cues rather than merely serving as a collection of easy single-stream samples.

\begin{table}[tb]
\centering
\caption{Comparison with single-modality reference baselines on the 210-sample aligned benchmark.}
\label{tab:single_modal_comparison}
\begin{tabular}{llccc}
\toprule
Modality & Method & ACC & UF1 & UAR \\
\midrule
\multirow{4}{*}{RGB}
& MMNET~\cite{li2022mmnet} & 0.4920 & 0.2509 & 0.2409 \\
&BDCNN~\cite{chen2022block}  & 0.4976 & 0.1329 & 0.1329 \\
&HTNet~\cite{wang2024htnet}  & 0.3667 & 0.1872 & 0.1944 \\
& The Released Benchmark & \textbf{0.5762} & \textbf{0.3507} & \textbf{0.3443} \\
\midrule
\multirow{3}{*}{Event}
& Voxel + CNN & 0.3667 & 0.3232 & 0.3093 \\
& Event Frame + CNN & 0.3524 & 0.2486 & 0.2301 \\
& The Released Benchmark & \textbf{0.5619} & \textbf{0.3180} & \textbf{0.3167} \\
\bottomrule
\end{tabular}
\end{table}

\section{Performance Analysis on MaE and ME Subsets}

\begin{table}[tb]
\centering
\caption{Performance comparison on the aligned benchmark under different expression subsets. The results in this table are re-obtained from exported sample-level predictions for subset-specific analysis, and may therefore differ slightly from the aggregate values reported in the main paper.}
\label{tab:macro_micro_comparison}
\begin{tabular}{llccc}
\toprule
Subset & Method & ACC & UF1 & UAR \\
\midrule
\multirow{3}{*}{Overall}
& RGB only & 0.5667 & 0.3399 & 0.3219 \\
& Event only & 0.5619 & 0.3006 & 0.3105 \\
& Ours & \textbf{0.6619} & \textbf{0.4912} & \textbf{0.4463} \\
\midrule
\multirow{3}{*}{MaE}
& RGB only & 0.5591 & 0.3315 & 0.3229 \\
& Event only & 0.5054 & 0.2946 & 0.3211 \\
& Ours & \textbf{0.6452} & \textbf{0.5033} & \textbf{0.4908} \\
\midrule
\multirow{3}{*}{ME}
& RGB only & 0.5726 & 0.3158 & 0.3309 \\
& Event only & 0.6068 & 0.4175 & 0.3988 \\
& Ours & \textbf{0.6752} & \textbf{0.5427} & \textbf{0.5032} \\
\bottomrule
\end{tabular}
\end{table}

\begin{figure}[tb]
  \centering
  \begin{subfigure}{0.3\linewidth}
    \includegraphics[width = \textwidth]{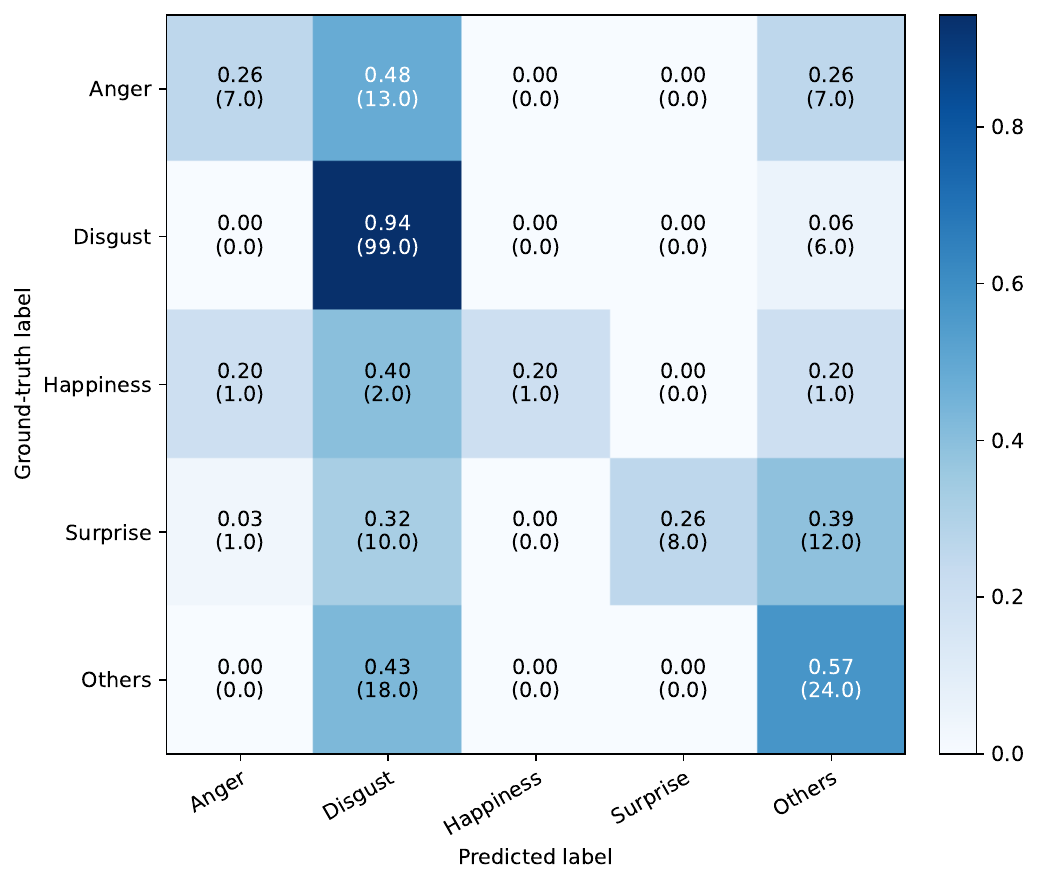}
    \caption{All 210 samples}
    \label{fig:all_cm}
  \end{subfigure}
  \hfill
  \begin{subfigure}{0.3\linewidth}
  \includegraphics[width = \textwidth]{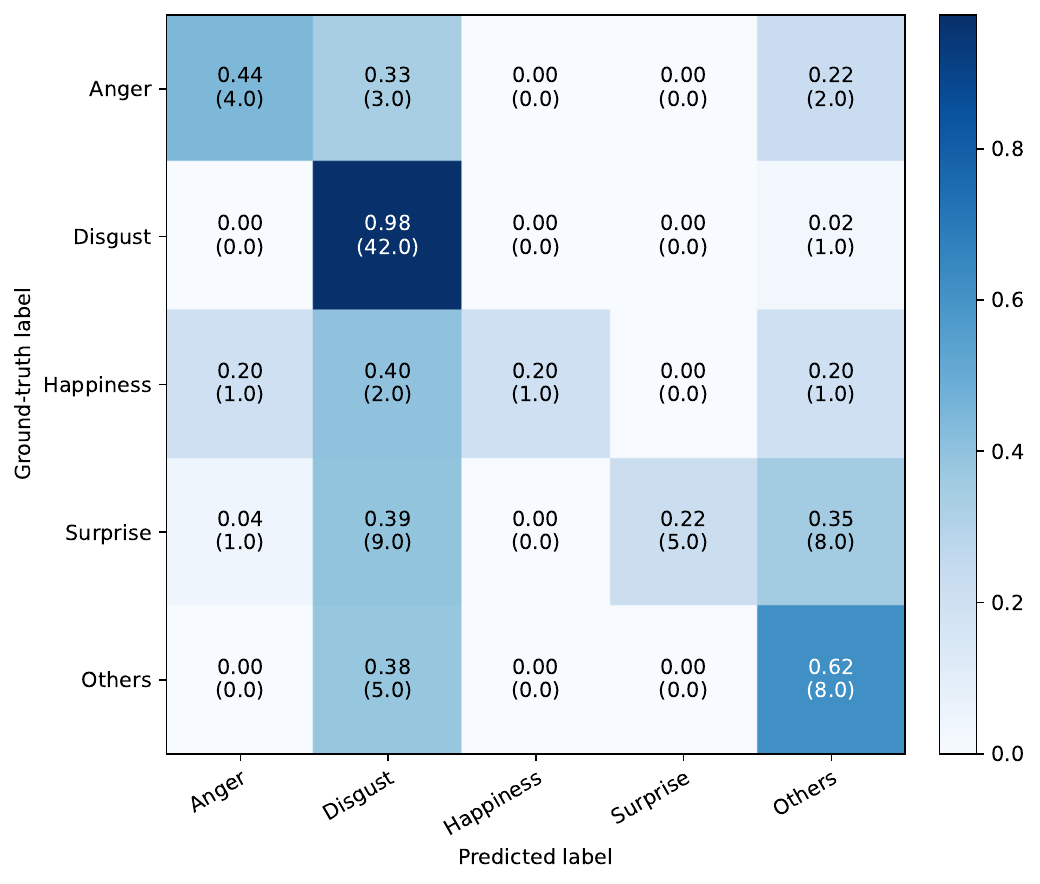}
    \caption{93 MaEs}
    \label{fig:mae_cm}
  \end{subfigure}
  \hfill
  \begin{subfigure}{0.3\linewidth}
   \includegraphics[width = \textwidth]{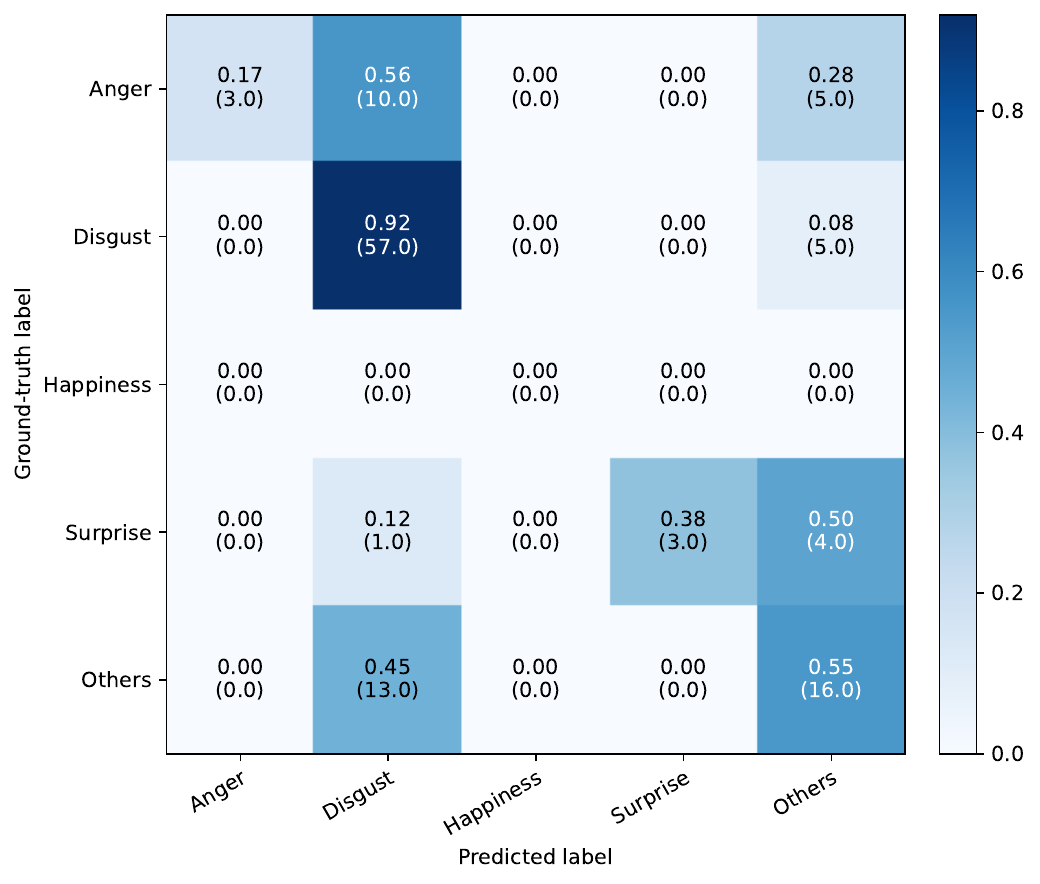}
    \caption{117 MEs}
    \label{fig:me_cm}
  \end{subfigure}
 \caption{Confusion matrices of the released multimodal reference baseline on the overall set, MaE subset, and  ME subset.}
\label{fig:confusion_matrix}

\end{figure}
To further analyze the behavior of the released benchmark beyond the aggregate setting, we additionally report results on the aligned subset after separating MaEs and MEs. This analysis is intended to reveal whether the relative advantages of multimodal learning remain stable across expression scales, and to provide a more fine-grained understanding of model behavior on CDER-SME.

\noindent\textbf{Note on result consistency.}
The results reported in this section are not numerically identical to the overall ablation values in the main paper. The reason is that, in the earlier stage of experimentation, we did not retain per-sample prediction outputs, whereas the present analysis requires sample-level predictions to separately evaluate MaEs and MEs. We therefore re-ran the corresponding experiments and exported the required predictions for subset-specific analysis. Due to the stochastic nature of network training, minor fluctuations in the aggregate metrics are expected. Nevertheless, the overall performance trend remains consistent: the multimodal setting still outperforms the RGB-only and event-only counterparts across all evaluation subsets.

Tab.~\ref{tab:macro_micro_comparison} summarizes the results. On the full aligned benchmark, the RGB-only baseline achieves an ACC/UF1/UAR of 0.5667/0.3399/0.3219, while the event-only baseline obtains 0.5619/0.3006/0.3105. In comparison, the released benchmark reaches 0.6619/0.4912/0.4463, showing a clear overall advantage over both unimodal counterparts. This indicates that the benchmark preserves meaningful complementarity between frame-based appearance cues and event-based transient motion signals.

On the MaE subset, the multimodal baseline continues to provide the strongest performance, achieving an ACC of 0.6452, a UF1 of 0.5033, and a UAR of 0.4908. These values are substantially higher than those of RGB only (0.5591/0.3315/0.3229) and event only (0.5054/0.2946/0.3211). This suggests that, even for relatively stronger facial motions, the combination of RGB and event modalities remains beneficial, likely because the two streams provide complementary structural and temporal information.

A similar conclusion holds for the ME subset. RGB only achieves 0.5726/0.3158/0.3309 in ACC/UF1/UAR, while event only reaches 0.6068/0.4175/0.3988. the released benchmark further improves the performance to 0.6752/0.5427/0.5032, again yielding the best results among the compared settings. Notably, the event-only branch is more competitive on MEs than on MaEs, which is consistent with the high temporal sensitivity of event sensing for capturing subtle and short-lived facial dynamics. However, the best performance is still achieved when event cues are combined with RGB information, indicating that high-frequency motion signals and stable appearance structure are jointly important for robust MER.

To further illustrate class-level behavior, we visualize the confusion matrices of the released benchmark for the overall set, the MaE subset, and the ME subset. Across all three settings, the largest confusion remains concentrated around visually similar or highly imbalanced categories, especially those involving \textit{Disgust}, \textit{Surprise}, and \textit{Others}. Nevertheless, the multimodal model preserves relatively strong diagonal responses in the dominant classes and maintains more balanced recognition behavior than the unimodal baselines, which is consistent with the quantitative improvements in UF1 and UAR.

Overall, the subset analysis supports two conclusions. First, the superiority of the multimodal baseline is stable across both MaE and ME settings, rather than being restricted to only one expression scale. Second, the stronger event-only competitiveness observed on the ME subset further validates the motivation of CDER-SME: event streams provide highly relevant temporal evidence for subtle facial dynamics, while RGB frames contribute complementary appearance context, making the combined benchmark particularly suitable for multimodal affective analysis.

\section{Hyperparameter Sensitivity Analysis} \label{subsec:hyperPara}
We further investigate the trade-off between the emotion classification objective and cross-modal similarity learning by varying the weights $\alpha$ (Cross-Entropy loss) and $\beta$ (Covariance loss). These analyses are included only to support reproducibility of the released reference baseline.

\begin{table}[tb]
\centering
\caption{Hyperparameter sensitivity analysis of the loss function weights. $\alpha$ and $\beta$ represent the weights for classification and modality similarity, respectively. Optimal results are \textbf{bolded}.}\label{tab:hyper_params}
\begin{tabular}{@{}ccccc@{}}
\toprule
\textbf{CE Weight ($\alpha$)} & \textbf{Cov Weight ($\beta$)} & \textbf{ACC} & \textbf{UF1} & \textbf{UAR} \\ \midrule0.0 & 1.0 & 0.5000 & 0.3577 & 0.3706 \\
0.2 & 0.8 & 0.6238 & 0.3706 & 0.3736 \\
0.4 & 0.6 & 0.6143 & 0.3983 & 0.3826 \\
0.6 & 0.4 & 0.6333 & 0.3744 & 0.3807 \\
\textbf{0.8} & \textbf{0.2} & \textbf{0.6667} & \textbf{0.4403} & \textbf{0.4246} \\
1.0 & 0.0 & 0.5904 & 0.3628 & 0.3536 \\ \bottomrule
\end{tabular}
\end{table}

The sensitivity analysis in Tab.~\ref{tab:hyper_params} reveals that model performance is highly dependent on the balance of these tasks. Relying exclusively on modality alignment ($\alpha=0.0$) or classification supervision ($\alpha=1.0$) results in suboptimal UF1 and UAR. The peak performance is observed at \textbf{$\alpha=0.8$ and $\beta=0.2$}, suggesting that while task-specific supervision is primary, a moderate level of covariance-based regularization is essential to structure the joint latent space effectively.

%
%

\end{document}